\documentclass[10pt,twocolumn,letterpaper]{article}

\usepackage[pagenumbers]{wacv} 

\usepackage{multirow}

\definecolor{wacvblue}{rgb}{0.21,0.49,0.74}
\usepackage[pagebackref,breaklinks,colorlinks,allcolors=wacvblue]{hyperref}

\def\wacvPaperID{401} 
\def\confName{WACV}
\def\confYear{2026}

\title{Improved Wildfire Spread Prediction with Time-Series Data and the WSTS+ Benchmark}

\author{Saad Lahrichi\\
University of Missouri\\
Columbia, MO 65201\\
{\tt\small saad.lahrichi@missouri.edu}
\and
Jake Bova\\
University of Montana\\
Missoula, MT 59802\\
{\tt\small jacob.bova@umt.edu}
\and
Jesse Johnson\\
University of Montana\\
Missoula, MT 59802\\
{\tt\small jesse.johnson@umt.edu}
\and
Jordan Malof\\
University of Missouri\\
Columbia, MO 65201\\
{\tt\small jmdrp@missouri.edu}
}

\begin{document}
\maketitle
\begin{abstract}
Recent research has demonstrated the potential of deep neural networks (DNNs) to accurately predict wildfire spread on a given day based upon high-dimensional explanatory data from a single preceding day, or from a time series of $T$ preceding days. For the first time, we investigate a large number of existing data-driven wildfire modeling strategies under controlled conditions, revealing the best modeling strategies and resulting in models that achieve state-of-the-art (SOTA) accuracy for both single-day and multi-day input scenarios, as evaluated on a large public benchmark for next-day wildfire spread, termed the WildfireSpreadTS (WSTS) benchmark. Consistent with prior work, we found that models using time-series input obtained the best overall accuracy, suggesting this is an important future area of research. Furthermore, we create a new benchmark, WSTS+, by incorporating four additional years of historical wildfire data into the WSTS benchmark.  Our benchmark doubles the number of unique years of historical data, expands its geographic scope, and, to our knowledge, represents the largest public benchmark for \textit{time-series}-based wildfire spread prediction.  
\end{abstract}       

\section{Introduction}
Wildfires are a global cause of concern that have severe human, economical, and environmental impacts, with the average annual economic burden from wildfires falling between \$71.1 billion and \$347.8 billion \cite{thomas2017costs}. In order to better manage, mitigate, and prevent wildfires, accurately predicting their spread is essential. In this work, we focus on the problem of next-day wildfire spread prediction, where we are provided with current and/or historical information about a particular wildfire, and then tasked with predicting its spatial extent on the following day.  

\begin{figure}
    \centering
    \includegraphics[width=0.9\linewidth]{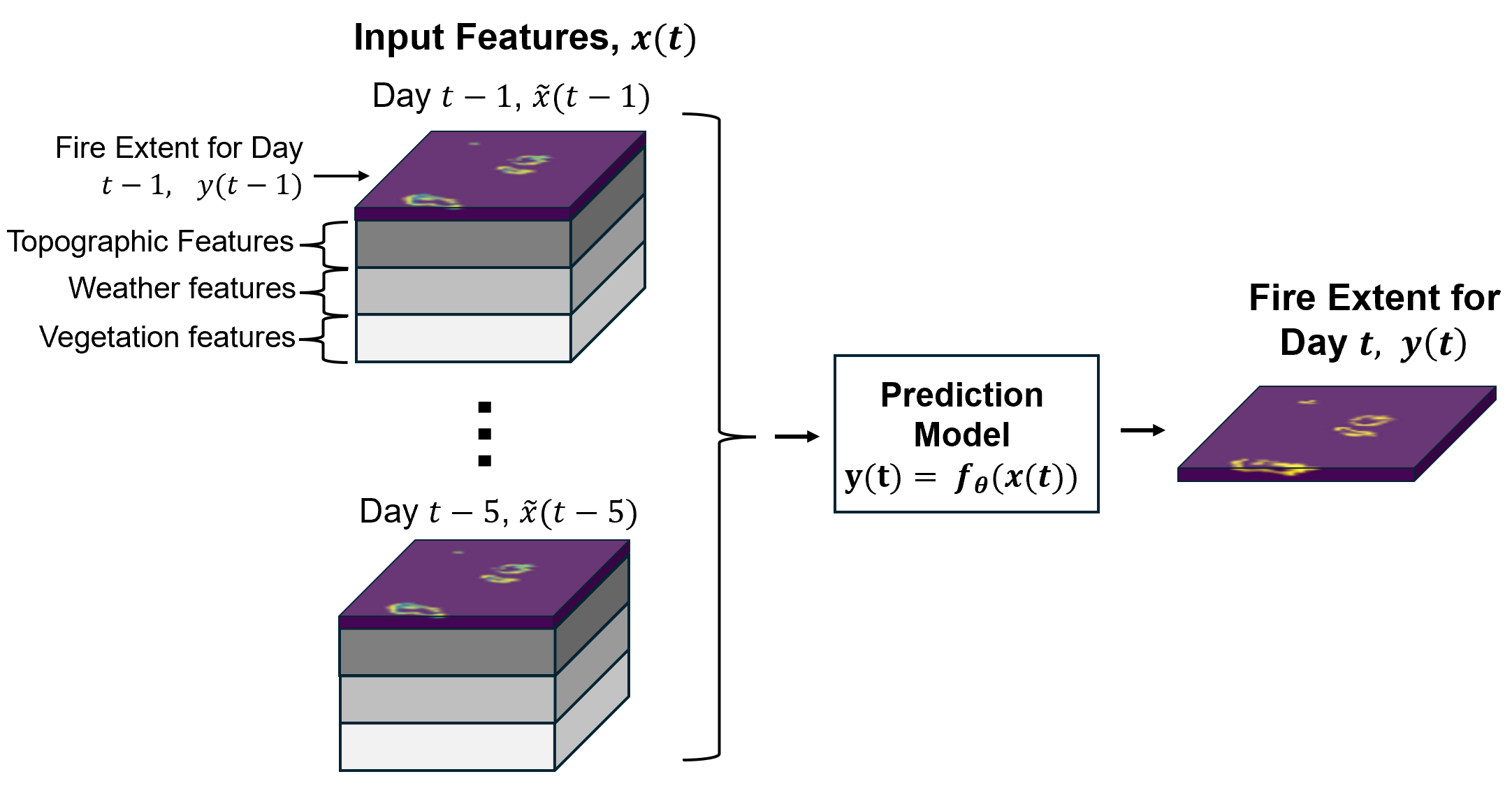}
    \caption{The wildfire prediction models take as input a geospatial map of several variables: vegetation, topography, and weather features, alongside the current day fire mask. We consider two scenarios: one in which the model receives input features from one preceding day, denoted $t-1$, and one in which it receives input from five previous days}
    \label{fig:problem_setting}
\end{figure}

A variety of approaches have been investigated to solve this problem, such as those based upon machine learning models \cite{khanmohammadi2022prediction, bot2022systematic, chetehouna2015predicting}, or physics-based and observationally-informed models \cite{Finney_1998, alexander2006evaluating, finney2006overview}.  In this work, however, we focus on a promising emerging class of techniques that utilize high-capacity machine learning models -- namely, deep neural networks (DNNs) -- to predict wildfire spread using high-dimensional explanatory input data.  These input data typically comprise a geospatial raster of the current extent of the fire, as well as explanatory features such as topography, climate, weather, and vegetation indices. Based upon these input data, the model is tasked with producing a geospatial map, or an image, reflecting the spatial extent of the fire on the following day. See \cref{fig:problem_setting} for an illustration.  

A variety of DNN-based models have been proposed to solve next-day prediction, including convolutional models \cite{liu2024fire, burge2023recurrent, marjani2024cnn}, attention-based models such as transformers \cite{shah2023wildfire}, and spatio-temporal models \cite{michail2024seasonal, bolt2022spatio}.  One major limitation of most existing work is the lack of standardized evaluation wherein studies often utilize different datasets, model training and evaluation procedures, or compare to few other existing methods. Furthermore, most existing research has focused on next-day prediction where only explanatory data from the current day is input (e.g.,  only $\tilde{x}(t-1)$ in \cref{fig:problem_setting}).  However, recent research found that models utilizing a time-series of $T$ previous days of data can achieve greater prediction accuracy \cite{gerard2023wildfirespreadts}, suggesting this as an important new direction in next-day wildfire prediction.  

\paragraph{Contributions of this Work} Our primary contributions are not a novel architecture, but rather, \textbf{our first contribution} is to perform a rigorous and controlled comparison of many existing approaches, both for single-day ($T=1$) and the time-series inputs ($T=5$).  We also propose a novel temporal positional embedding.  We compare all methods using the same public benchmark dataset, along with the same training, hyperparameter optimization, and evaluation procedures.  We perform all experiments on the WildfireSpreadTS (WSTS) benchmark \cite{gerard2023wildfirespreadts} because it is the only public benchmark for time-series wildfire prediction, it sufficiently large to support DNN training and evaluation, and in contrast to all other existing benchmarks, it employs a realistic 12-fold \textit{leave-one-year-out} cross-validation. Our study reveals the best existing modeling strategies, resulting in substantial performance improvements over the current state-of-the-art (SOTA) for the WSTS benchmark.  

\textbf{Our second contribution} is to introduce WSTS+, an extended benchmark for next-day wildfire spread, constructed by doubling the number of years of historical wildfire events in WSTS.  By adopting WSTS+, we can double the size of our training datasets. We conduct experiments that reveal additional historical training data, either in WSTS or WSTS+, yields little improvement in the accuracy of wildfire models. We discover and analyze significant cross-year domain shift, a critical challenge for the field.

The rest of the paper is structured as follows: we formulate our problem setting in \cref{sec:problem_setting}, \cref{sec:related_works} reviews related works, \cref{sec:wsts} describes WSTS, \cref{sec:experiments} details our experiments, \cref{sec:wsts+} introduces WSTS+, and \cref{sec:conclusion} concludes.

\section{Problem Setting}
\label{sec:problem_setting}
In its general formulation, the goal of next-day wildfire spread prediction is to predict a wildfire's spatial extent on some $t^{th}$ day, denoted $y(t)$, given explanatory data from one or more \textit{preceding} days, denoted $x(t)$.  We adopt the more specific settings of recent literature \cite{gerard2023wildfirespreadts,huot2022next}, as illustrated in \cref{fig:problem_setting}, which assume there are $T$ consecutive previous days of explanatory data, so that $x(t)=\{ \tilde{x}(t-i) \}_{i=1}^{T}$, and each $\tilde{x}(t)$ comprises a geospatial raster, so that $\tilde{x}(t) \in \mathbb{R}^{H \times W \times C}$, where $H, W$ correspond to spatial dimensions, and $C$ represents the number of explanatory variables, which may include previous fire masks (e.g., $y(t) \subset \tilde{x}(t)$).  The fire extent is encoded in a binary geospatial image, $y(t) \in  \{0,1\}^{H \times W}$, where a value of one indicates the presence of a fire. Our goal is then to use a dataset of historical wildfire data to infer parameters, $\theta$, of a predictive model of the form $y(t) = f_{\theta}(x(t))$.    

\section{Related Works}
\label{sec:related_works}
\paragraph{Next Day Wildfire Segmentation} DNN-based segmentation has received growing attention due to its accuracy, enabled by the recent development of large datasets of historical fire data.  \cite{huot2022next} created Next-Day Wildfire Spread, a large and public dataset for \textit{next-day} spread prediction, and used it to train a custom deep segmentation model. Concurrently, \cite{prapas2022deep} developed SeasFire Cube and trained Unet++ models \cite{zhou2018unet++} for medium-term fire prediction, between 8 and 64 days. \cite{kondylatos2022wildfire} improved upon the collected data cube and found that the LSTM and ConvLSTM models outperformed the Fire Weather Index (FWI). In FireSight, \cite{gottfriedsenfiresight} collected a dataset using remote sensing data from 20 datasets, and trained a 3D UNet model to model short-term fire hazard, between 3 and 8 days.  Overall, most work has been done using U-Net architectures and their variants, and many authors \cite{li2024wildfire, fitzgerald2023paying, shah2023wildfire, xiao2024wildfire} have recently reported that attention-based U-Nets achieve greater accuracy.  We investigate the SwinUnet\cite{liu2021swin} in our study as a widely used and therefore representative example of such models. 

\paragraph{Next Day Wildfire Prediction with Time-Series} In contrast to the existing work discussed above, we also focus on time-series input for spread prediction, which has been cited as an important emerging direction \cite{fitzgerald2023paying, gerard2023wildfirespreadts, li2024wildfire}.  Historically, time-series modeling has been challenging due to the lack of appropriate public datasets to train and evaluate models for this task. Recently, \cite{gerard2023wildfirespreadts} extended the Next-Day Wildfire Spread dataset from \cite{huot2022next} to be suitable for time-series prediction, and achieved their best overall next-day predictions using a time-series model, termed UTAE \cite{garnot2021panoptic}. 

\paragraph{Other DL Approaches}  Aside from a segmentation formulation, researchers have also investigated, for example, reinforcement learning \cite{ross2021being}, probabilistic cellular automata \cite{ghosh2024fire}, and synthetic data approaches \cite{lisim2real}. We refer readers to \cite{xu2024wildfire} for a review of DL for wildfire prediction.

\section{The WSTS Benchmark}
\label{sec:wsts}
In this work, we employ the WildfireSpreadTS benchmark \cite{gerard2023wildfirespreadts}. The dataset includes 607 wildfire events across the western United States between 2018 and 2021, totaling 13,607 daily multi-channel images. These 23 channels include data on active fires, weather, topography, and vegetation, resampled to a common resolution of 375 meters, providing a multi-modal and multi-temporal framework for modeling fire spread. A key feature of this benchmark is a rigorous 12-fold cross-validation evaluation procedure. Each fold of the cross-validation includes all wildfire events from a single year, so that the trained models are always evaluated on wildfire events from a previously unseen year, reflecting real-world use of wildfire prediction models.   
\section{Improving Wildfire Spread Prediction}
\label{sec:experiments}

In this section, we describe our methods for $T=1$ and $T=5$ scenarios, respectively, as well as experiments to support them (e.g., ablations). Results for our developed benchmark models are reported in \cref{tab:results}, in terms of Average Precision (AP) using 12-fold leave-one-year-out cross-validation on the WSTS benchmark, following prior work \cite{gerard2023wildfirespreadts}. Also following \cite{gerard2023wildfirespreadts}, we report model performance for three feature sets: vegetation features only (Veg), a combination of vegetation and topographic features (Multi), and all features (All), which includes additional weather forecast features.  Detailed descriptions of our models, alongside full experimental details, are provided in \cref{app:exp_details} in the supplement. Key deployment information (FLOPs, GPU memory requirements, training/inference times) can be found in \cref{sub:deployment_info}.  Models in \cref{tab:results} with citations correspond to the three current best models on WSTS, as reported in \cite{gerard2023wildfirespreadts}. All other models reported in \cref{tab:results} were developed in this work. 


\subsection{Single-Day Input ($T=1$)}
\label{sec:results_single_day_input}

The current $T=1$ SOTA utilizes a U-Net architecture with a ResNet-18 encoder, and is denoted Res18-Unet\cite{gerard2023wildfirespreadts}. Therefore, we focus our investigation on improving the Res18-Unet\cite{gerard2023wildfirespreadts}.     

\paragraph{Modeling Improvements} We next describe the investigated improvements to the Res18-Unet\cite{gerard2023wildfirespreadts} at a high level.  More details can be found in \cref{app:exp_details} of the supplement. 

\textit{(i) Encoders.} Better performance may be obtained with larger encoders or those with attention mechanisms. Recent studies have indicated that attention-based models may be superior to convolutional models for wildfire spread \cite{li2024wildfire, zou2023attention, shah2023wildfire, xiao2024wildfire}.  We investigate a ResNet50 \cite{he2016deep} encoder, as well as the attention-based SwinUnet-Tiny encoder \cite{cao2022swin} and SegFormer-B2 \cite{xie2021segformer}.  

\textit{(ii) Utilizing Pre-trained Parameters.} Utilizing pre-trained weights to initialize training is a well-established technique to improve model accuracy, including in remote sensing \cite{iglovikov2018ternausnet}. We investigate pre-trained weights for each of the encoders that we consider (i.e., ResNet18, ResNet50, SwinUnet, and SegFormer), while the decoders are trained from scratch.  

\textit{(iii) Improved Loss Functions.} The existing SOTA Res18-Unet \cite{gerard2023wildfirespreadts} is trained using weighted binary cross-entropy loss. However, it has been established that Jaccard/Dice losses are often superior alternatives for segmentation tasks \cite{eelbode2020optimization}, and focal loss has been shown effective for class imbalance \cite{lin2017focal} (the WSTS benchmark exhibits severe class imbalance), and for wildfire spread in particular \cite{fitzgerald2023paying}. Therefore we investigate and compare the aforementioned losses in our experiments.      

\textit{(iv) Improved Hyperparameter Optimization.} The existing SOTA Res18-Unet \cite{gerard2023wildfirespreadts} was trained by selecting the model with the highest F1 score on the validation; however, all models on WSTS are evaluated utilizing the average precision (AP) metric \cite{gerard2023wildfirespreadts}.  We investigate aligning the validation and testing metrics by using AP for both.    

For our experiments, we consider a U-Net with a ResNet-18 encoder (denoted \textit{Res18-Unet}), a ResNet-50 encoder (denoted \textit{Res50-Unet}), a SwinUnet-Tiny (denoted \textit{SwinUnet}), and a SegFormer-B2 (denoted \textit{SegFormer}).  For each of these models, we perform a grid search over all combinations of learning rates ($[1e-1,1e-2,1e-3,1e-4, 1e-5]$), loss functions (BCE, Focal, Dice, Jaccard), and the use of pre-training or not (a binary choice).  Following \cite{gerard2023wildfirespreadts}, we use a single fold of the 12-fold cross-validation, and only one of the three feature sets (the "All" set) for this optimization. As discussed, in contrast to previous work, we utilize AP during validation to select the best models instead of F1.  The focal loss has two hyperparameters: $\alpha$, set as the inverse frequency of positive class pixels, and $\gamma$, set to its default value of two.  

\paragraph{Experimental Results} We found that pre-training was nearly always beneficial, and that Focal Loss usually yielded substantial improvements compared to our other candidate losses. Therefore, for the WSTS benchmark, we included both pre-training and focal loss in all our models: \textit{Res18-Unet}, \textit{Res50-Unet}, \textit{SwinUnet}, and \textit{SegFormer}. As an ablation study, \cref{tab:ablation_t1} reports the performance of our \textit{Res18-Unet} on the full WSTS benchmark, where we progressively remove each of our improvements to assess its impact. Our results indicate that each improvement is highly beneficial, or at least not significantly harmful.     

\cref{tab:results} reports the performance of our models on the WSTS benchmark, compared to the best existing $T=1$ model, \textit{Res18-Unet\cite{gerard2023wildfirespreadts}}. Our \textit{Res18-Unet} is identical to the \textit{Res18-Unet\cite{gerard2023wildfirespreadts}}, except for our aforementioned modifications, and obtains substantially higher AP across all input features considered: a 37\% improvement on average. We find that these improvements are statistically significant using a Wilcoxon signed-rank test, and report the results in \cref{sub:stat_significance}. 

Our other models, \textit{Res50-Unet}, \textit{SwinUnet}, and \textit{SegFormer} also substantially outperform the existing \textit{Res18-Unet\cite{gerard2023wildfirespreadts}}. However, despite having approximately twice the number of trainable model parameters of \textit{Res18-Unet}, we find that our model outperforms the three larger models in most cases. \textit{Our Res18-Unet also obtains the highest overall AP (0.468) for the $T=1$ models when utilizing the "Multi" feature set, establishing a new SOTA on WSTS for $T=1$}.  

Several recent studies have reported that large and/or attention-based models achieve SOTA accuracy for $T=1$ wildfire spread prediction \cite{li2024wildfire, zou2023attention, shah2023wildfire, xiao2024wildfire}.  However, we find here that with simple improvements and appropriate optimization, \textit{Res18-Unet} outperforms such models. In \cref{sub:simpler_outperform_complex} and \cref{tab:loyo_vs_random}, we test the hypothesis that the more rigorous (and potentially more real-world) leave-one-year-out cross-validation adopted by the WSTS benchmark may penalize more complex models for overfitting. Yet, we find that using a random cross-validation still allows the Res18-Unet to outperform attention-based models, suggesting its superiority even under different evaluation scenarios.

\begin{figure}
    \centering
    \includegraphics[width=\linewidth]{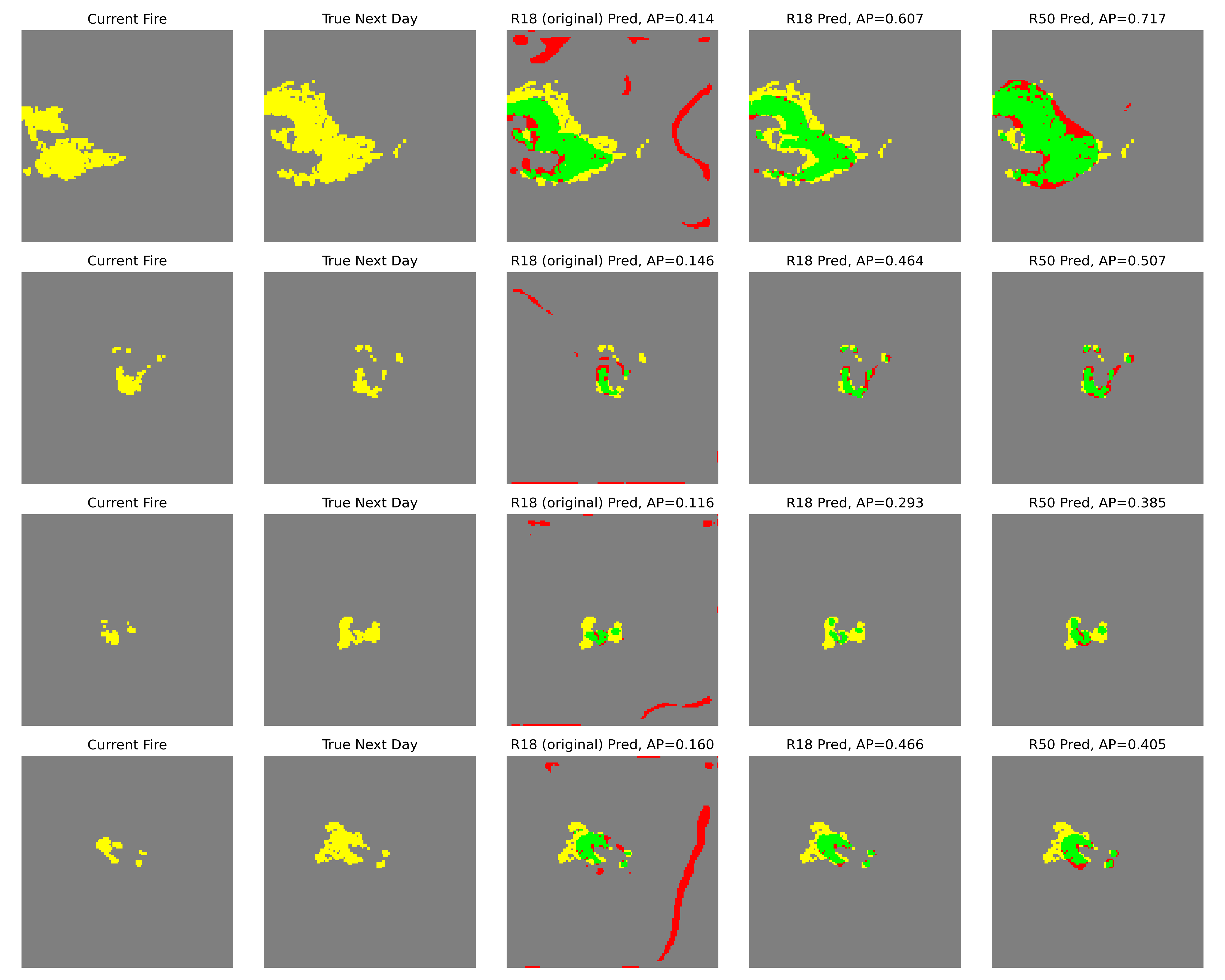}
    \caption{Sample predictions made by the Res18-Unet \cite{gerard2023wildfirespreadts}, our \textit{Res18-Unet}, and \textit{Res50-Unet}. The two leftmost columns show the current fire spread $y(t-1)$ and the next-day label $y(t)$. True positive pixels are colored in green, while false positives are colored in red}
    \label{fig:example_predictions}
\end{figure}

In \cref{fig:example_predictions}, we qualitatively evaluate our \textit{Res18-Unet} and \textit{Res50-Unet} against the Res18-Unet \cite{gerard2023wildfirespreadts}. Each row corresponds to a fire event, and the columns show the current fire, the next-day label, and the predictions of each model. Yellow represents the fire extent, green shows correctly predicted burned areas, and red shows false positives. 

We observe that the original model tends to overpredict fire spread, leading to multiple red patches where no fire actually occurs. However, the model also underpredicts in areas where the fire spreads, capturing some, but not the full extent of the fire. On the other hand, we observe that our models make consistently more accurate predictions, with far fewer false positives, and slightly better matching green areas. 

We provide additional examples in \cref{fig:success_2018}, through \cref{fig:failure_2021} of the best and worst predictions made by the Res18-Unet for each testing year, and we analyze these results in \cref{sub:failure_cases} and \cref{sub:fire_size_impact}. Overall, we find that the model struggles with very small, newly ignited, or displaced fires, while achieving high accuracy on larger, more consolidated fires, with performance positively correlated with fire size across years.

\begin{table}[]
\caption{Ablation showing the impact of the successive removal of each of our improvements on a \textit{Res18-Unet} trained on Vegetation features}
    \centering
    \begin{tabular}{lcc}
        \toprule
        \textbf{Model} & \textbf{Test AP} & \textbf{Percent Decrease} \\
        \midrule
        Res18-Unet (ours) & $0.455 \pm 0.092$ & $-$ \\
        No pretraining & $0.456 \pm 0.086$ & $-0.22$ \\
        No focal loss & $0.345 \pm 0.084$ & $24.18$ \\
        No AP as validation & $0.321 \pm 0.078$ & $29.45$ \\
        \bottomrule
    \end{tabular}
    
    \label{tab:ablation_t1}
\end{table}

\begin{table*}[h]
    \centering
    \caption{Mean test AP $\pm$ standard deviation using vegetation features only (Veg), vegetation, land cover, topography and weather (Multi) and All features, when training with 1 and 5 input days. Models with citations represent accuracy reported on our benchmark from previous publications; all other models reported are developed in this work.  Results style: \textbf{best} }
    \begin{tabular}{clcccccc}
        \toprule
        {\textbf{Fusion Level}} & \textbf{Model} & \textbf{Input days} & \textbf{Veg} & \textbf{Multi} & \textbf{All} & \textbf{\# Params} \\
        \midrule
        \multirow{4}{*}{-} & Res18-Unet\cite{gerard2023wildfirespreadts} & 1 & $0.328 \pm 0.090$ & $0.341 \pm 0.085$ & $0.341 \pm 0.086$ & 14.3M \\
                            & Res18-Unet & 1 & ${0.455 \pm 0.090}$ & $\mathbf{0.468 \pm 0.087}$ & $\mathbf{0.460 \pm 0.084}$ & 14.3M \\
                            & Res50-Unet & 1 & $\mathbf{0.457 \pm 0.089}$ & ${0.459 \pm 0.090}$ & ${0.451 \pm 0.093}$ & 32.5M \\
                            & SwinUnet & 1 & $0.432 \pm 0.088$ & $0.437 \pm 0.082$ & $0.424 \pm 0.090$ & 27.2M \\
                            & SegFormer & 1 & $0.433 \pm 0.080$ & $0.436 \pm 0.083$ & $0.423 \pm 0.087$ & 27.5M \\
        \cmidrule{2-7} 
        \multirow{3}{*}{Data} & Res18-Unet\cite{gerard2023wildfirespreadts} & 5 & $0.333 \pm 0.079$ & $0.344 \pm 0.076$ & ${0.325 \pm 0.108}$ & 14.4M \\
                              & Res18-Unet & 5 & $\mathbf{0.472 \pm 0.083}$ & $\mathbf{0.469 \pm 0.087}$ & $\mathbf{0.460 \pm 0.084}$ & 14.4M \\
                              &  SwinUnet & 5 & $0.447 \pm 0.087$ & $0.453 \pm 0.083$ & $0.435 \pm 0.079$ & 27.3M \\
                              & SegFormer & 5 & $0.439 \pm 0.081$ & $0.436 \pm 0.085$ & $0.430 \pm 0.082$ & 27.7M \\
        \cmidrule{2-7} 
        \multirow{4}{*}{Feature} 
                              &  UTAE\cite{gerard2023wildfirespreadts} & 5 & $0.372 \pm 0.088$ & $0.350 \pm 0.113$ & $0.321 \pm 0.135$ & 1.1M \\
                              &  UTAE & 5 & $0.452 \pm 0.082$ & $0.459 \pm 0.088$ & $0.433 \pm 0.099$ & 1.1M \\
                              &  UTAE(Res18) & 5 & $\mathbf{0.478 \pm 0.085}$ & $\mathbf{0.477 \pm 0.089}$ & $\mathbf{0.475 \pm 0.091}$ & 14.6M \\
                              
        \bottomrule
    \end{tabular}
    \label{tab:results}
\end{table*}

\subsection{Time-Series Input, $T=5$}

Existing models for the time-series scenario generally adopt one of two approaches: (i) a data-level fusion, or (ii) a feature-level fusion.  In data-level fusion, the features for each day, $\tilde{x}(t) \in \mathbb{R}^{H \times W \times C}$ are concatenated along the feature dimension into one input tensor, $x(t)= [ \tilde{x}(t-1) |, ... ,| \tilde{x}(t-T) ] \in \mathbb{R}^{H \times W \times CT}$, after which they can be processed in the same manner as single-day input (see \cref{sec:problem_setting} for problem notation).  Therefore, we adopt our best-performing $T=1$ models from \cref{sec:results_single_day_input}, and their hyperparameter settings, and evaluate them for data-level fusion.  As a reference, we also include the \textit{reported} results of the Res18-Unet \cite{gerard2023wildfirespreadts} when it was applied for data-level fusion.   

In this context, feature-level fusion implies that we use a shared encoder to first extract features (or embeddings) independently for each day of our input, $\tilde{z}(t) = f_{\theta_{En}}(\tilde{x}(t))$ so that we have a collection of features, $z(t)=\{ \tilde{z}(t-i) \}_{i=1}^{T}$, which are utilized as input into a subsequent model for joint processing (i.e., fusion).  The current SOTA accuracy on WSTS, both for the time-series setting, and overall, was obtained with a UTAE model \cite{garnot2021panoptic}, as reported in \cite{gerard2023wildfirespreadts}. Furthermore, the UTAE achieved superior accuracy despite having just 1.1M parameters - significantly fewer than many other models considered (e.g., the Res18-Unet has 14.3M). Therefore, we focus our modeling improvements on the UTAE from \cite{gerard2023wildfirespreadts}. 


\paragraph{Improvements to the UTAE} We develop two improved UTAE models, referred to as \textit{UTAE} and \textit{UTAE(Res18)}. \cref{sub:utae} introduces key notation and background supporting the design of each model below, and the new positional encodings.

\textit{Our UTAE Model.} Our \textit{UTAE} includes two major improvements over the \textit{UTAE\cite{gerard2023wildfirespreadts}}. The first improvement is to adopt all of the changes investigated for the single-day models from \cref{sec:results_single_day_input}. Pursuant to this, following previous work convention, we did a joint search over the following hyperparameters using a single fold of the WSTS benchmark: pre-training (or not), learning rates ($[1e-2,1e-3,1e-4,1e-5]$), and the type of loss (Focal, BCE, Jaccard, and Dice loss).  The second improvement is the introduction of a novel positional encoding in the temporal fusion utilized by the \textit{UTAE}. To our knowledge, this modification is novel within the vision and wildfire literature.  Specifically, instead of using day-of-year positional encodings, as done in \cite{garnot2021panoptic, gerard2023wildfirespreadts}, where $\bar{t} \in [1,365]$, we propose to use a absolute positional encoding that indicates the relative position of each day's set of features within the time-series, so that $\bar{t} \in [1,...,T]$ for a T-day input.  We hypothesize that the features (especially the fire mask) from the most recent day of the fire will be most important for making predictions, and therefore, this relative position information will be much more important than its position in the year. Furthermore, it may be difficult for the models to infer relative positional information from day-of-year encodings, potentially undermining performance.   

\textit{Our UTAE(Res18) Model.} This model is obtained by making one additional improvement to our \textit{UTAE} model.  The encoder utilized in the \textit{UTAE\cite{gerard2023wildfirespreadts}} is relatively small (in terms of free parameters).  Therefore, in a similar fashion to our investigation in \cref{sec:results_single_day_input}, we replace the existing UTAE's encoder with a pre-trained ResNet-18.  

\paragraph{Experimental Results} \cref{tab:results} reports the accuracy (in terms of AP) of our time-series models on the WSTS benchmark, categorized by the type of fusion performed: data-level or feature-level. Regarding data-level fusion, our \textit{Res18-Unet}, \textit{Swin-Unet}, and \textit{SegFormer} all substantially outperform the existing \textit{Res18-Unet\cite{gerard2023wildfirespreadts}} across all combinations of input features, with the \textit{Res18-Unet} providing the best overall AP (AP=0.472, on Vegetation features).  Regarding feature-level fusion, our two UTAE models (\textit{UTAE} and \textit{UTAE(Res18)}) substantially outperform the existing \textit{UTAE\cite{gerard2023wildfirespreadts}}, which is the current SOTA model on WSTS, both for time-series input ($T>1$) and overall.  Our \textit{UTAE(Res18)} model achieves the highest overall performance for each combination of input features, across both single-day and time-series models.  \textit{In particular, our UTAE(Res18) achieves the highest overall AP with the Vegetation (Veg) feature subset, leading to a new overall SOTA performance on WSTS of AP=0.478.}

Notably, our results indicate that models receiving time-series input generally outperform those with single-day input.  This is especially apparent when comparing data-level fusion models, such as \textit{Res18-Unet} and \textit{SwinUnet}, with their single-day counterparts, since they have few architectural differences.  Most existing wildfire spread prediction in the literature has focused on the single-day input; however, our findings here corroborate those from \cite{gerard2023wildfirespreadts} and suggest that time-series modeling is a promising emerging modeling strategy.  

Our results also provide evidence that each modeling change is beneficial. As discussed, our \textit{UTAE} included several applicable improvements discussed for our single-day models in \cref{sec:results_single_day_input}, as well as our improved temporal encodings described in this sub-section.  We therefore conducted an ablation experiment, reported in \cref{tab:ablation_pos_enc}, to demonstrate that our modified positional encodings provide additional benefits. To show that the pre-trained ResNet-18 encoder is beneficial, we can compare the performance of \textit{UTAE(Res18)} and \textit{UTAE} in \cref{tab:results}: the pre-trained ResNet-18 is the only difference between these two models.   

\begin{table}[]
    \centering
    \caption{Test AP of UTAE trained on Vegetation features using the original Absolute positional encodings from \cite{gerard2023wildfirespreadts}, versus our proposed Relative positional encodings}
    \begin{tabular}{lcc}
    \toprule
    \textbf{Pos. Encodings} &  \textbf{Absolute} & \textbf{Relative}\\ 
    \midrule
    UTAE & $0.419 \pm 0.101$ & $\mathbf{0.452 \pm 0.082}$\\
    \bottomrule
    \end{tabular}
    \label{tab:ablation_pos_enc}
\end{table}

Finally, we observe that increasing the number of input features is not always beneficial, which is consistent with \cite{gerard2023wildfirespreadts}, where the best AP was often achieved with the Veg or Multi feature sets rather than the All set. This suggests that the explanatory power of some features is outweighed by the cost of (often significantly) increasing the input dimensionality. We hypothesize that this may be due to the low resolution and/or noise present in some features, such as the weather forecast features, which have a resolution of 27 km, while the fire masks have a resolution of 375 m.

\section{The WSTS+ Benchmark}
\label{sec:wsts+}

\begin{figure}
    \centering
    \includegraphics[width=0.95\linewidth]{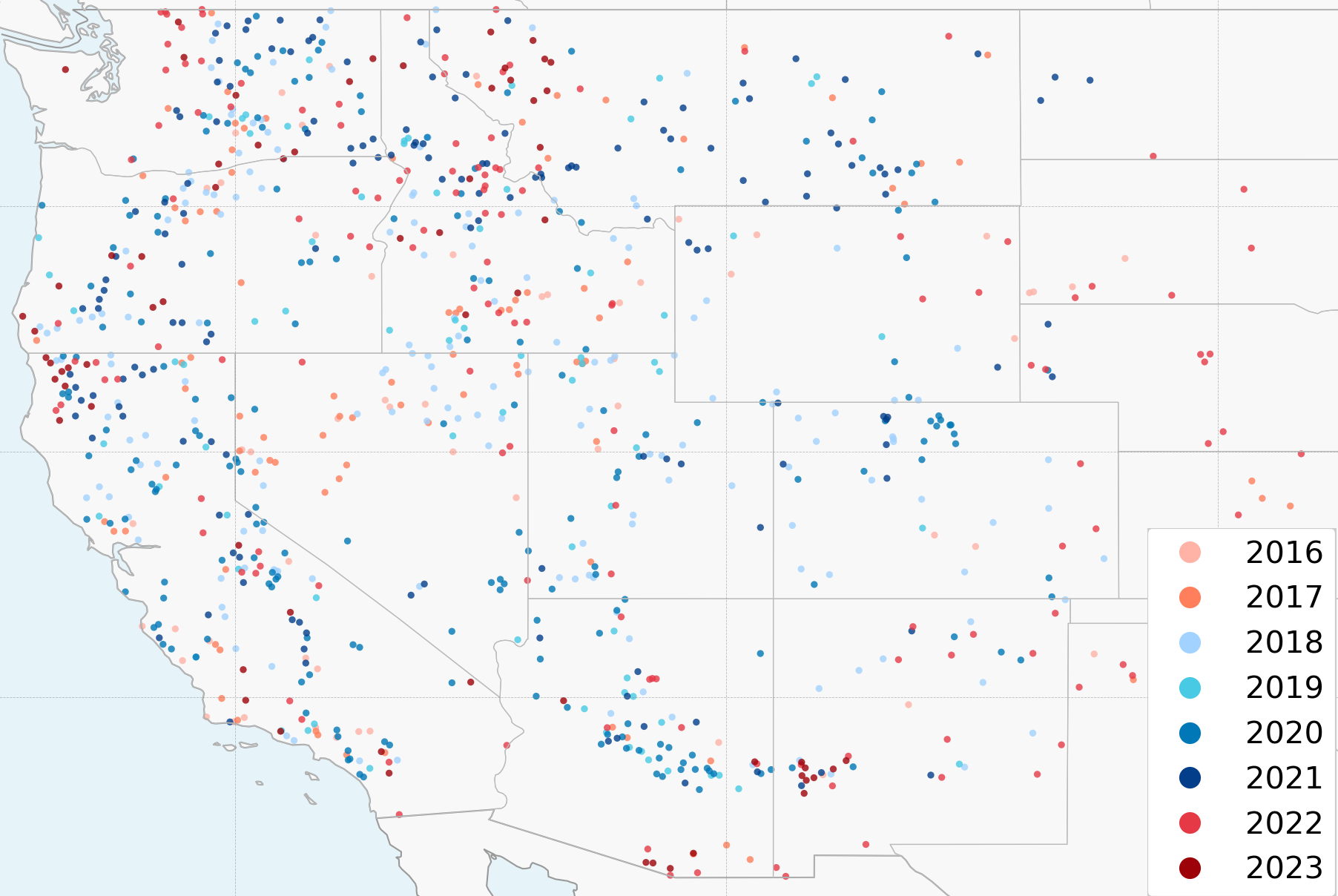}
    \caption{Geographic distribution of the fire events in each year of WSTS (blue) and WSTS+ (red)}
    \label{fig:wsts+map}
\end{figure}

Our results on the WSTS benchmark indicated that relatively simple models performed best, such as those based upon a ResNet-18, rather than models utilizing larger encoders (e.g., ResNet-50) or those utilizing attention (e.g., SwinUnet).  This contrasts sharply with the broader vision literature where larger models tend to perform best, given sufficient quantities of training data.  Therefore, we hypothesize that collecting more training data would facilitate the use of larger models, yielding superior modeling performance. To investigate this hypothesis, we expand the original WSTS benchmark by curating four additional years of historical wildfire data: 2016, 2017, 2022, and 2023. Our extended dataset, termed WSTS+, contains twice the number of years of historical wildfire data, expands the geographic diversity of the benchmark, and is -- to our knowledge -- the largest public benchmark for time-series next-day wildifre spread prediction. We visualize the geographic distribution of WSTS+ events in \cref{fig:wsts+map} and find that it much of the new data is in the Western United States, similar to WSTS, but that it includes some unique locations there, and some additional data in the eastern states. \cref{tab:wsts+} summarizes the differences between both datasets in terms of numbers of years, fire events, total images, and active fire pixels. Further collection details can be found in \cref{sub:collection_details} of the supplement.

\begin{table}[h!]
    \centering
    \caption{Comparison between the original WSTS dataset and our extension. We double the number of years and total images and drastically increase the number of fire events and active fire pixels.}
    \resizebox{\columnwidth}{!}{
    \begin{tabular}{lccc}
    \toprule
    \textbf{Dataset}        & \textbf{WSTS} & \textbf{WSTS+} & \textbf{Increase} \textbf{(\%)} \\
    \midrule
    Years     & 4 (2018-2021) & 8 (2016-2023) & $+100$ \\
    Fire Events & 607 & 1,005 & $+65.6$ \\
    Total Images & 13,607 & 24,462 & $+79.8$ \\
    Active Fire Px & 1,878,679 & 2,638,537 & $+40.4$ \\
    \bottomrule
    \end{tabular}
    }
    \label{tab:wsts+}

\end{table}

\subsection{Benchmarking Models with WSTS+}
\begin{figure}
    \centering
    \includegraphics[width=0.85\linewidth]{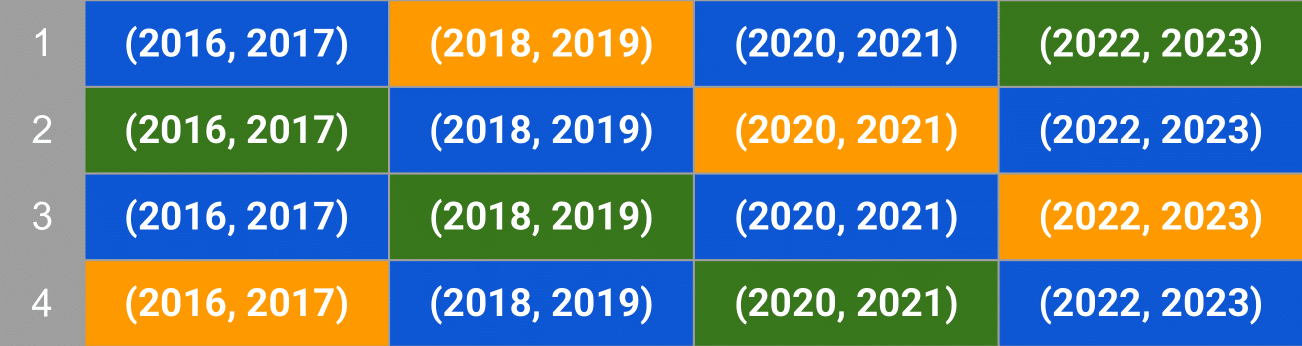}
    \caption{New cross-validation folds used for WSTS+. Each pair of consecutive years is used as validation/testing once. Color code: blue: training, orange: validation, green: test}
    \label{fig:wsts+folds}
\end{figure}

As compared to WSTS, we propose a new scheme for evaluating models using WSTS+, which exploits its greater size to significantly reduce computational complexity compared to WSTS's 12-fold cross-validation -- thereby making the benchmark more accessible to researchers -- while maintaining a similar level of real-world rigor.  For WSTS+, we propose to divide the available data into four folds that each contain two consecutive years of historical wildfire data.  We then evaluate models using four-fold cross-validation, where in each iteration, one fold of data is used for testing, one fold for validation, and two folds for training, as illustrated in \cref{fig:wsts+folds}. To ensure that the testing and validation sets have the same relative temporal distance to the training set, we always select them so that they are non-consecutive.  This results in four-fold cross-validation instead of the twelve-fold cross-validation utilized in WSTS, making it far less computationally intensive. At the same time, this approach doubles the quantity of data in the training and validation sets, ideally allowing researchers to train larger and more sophisticated models. Lastly, because two consecutive years of data are included in the test set, the benchmark still evaluates models under challenging realistic testing conditions.   

\begin{table}[h]
    \centering
    \caption{Mean test AP $\pm$ standard deviation using vegetation features only (Veg), vegetation, land cover, topography and weather (Multi) and All features, when training on the WSTS+ dataset Results style: \textbf{best} }
    \resizebox{\columnwidth}{!}{
    \begin{tabular}{lccccc}
        \toprule
        \textbf{Model} & \textbf{Days} & \textbf{Veg} & \textbf{Multi} & \textbf{All}  \\
        \midrule
        Res18-Unet & 1 & $0.349 \pm 0.109$ & $0.351 \pm 0.105$ & $\mathbf{0.351 \pm 0.122}$ \\
        Res50-Unet & 1 & ${0.345 \pm 0.096}$ & ${0.353 \pm 0.122}$ & $\mathbf{0.351 \pm 0.122}$\\
        UTAE(Res18) & 5 & $\mathbf{0.354 \pm 0.113}$ & $\mathbf{0.363 \pm 0.129}$ & ${0.350 \pm 0.117}$  \\
        \bottomrule
    \end{tabular}
    }
    \label{tab:WSTS+results}
\end{table}

\subsection{Experimental Results with WSTS+}
\begin{figure}
    \centering
    \includegraphics[width=0.9\linewidth]{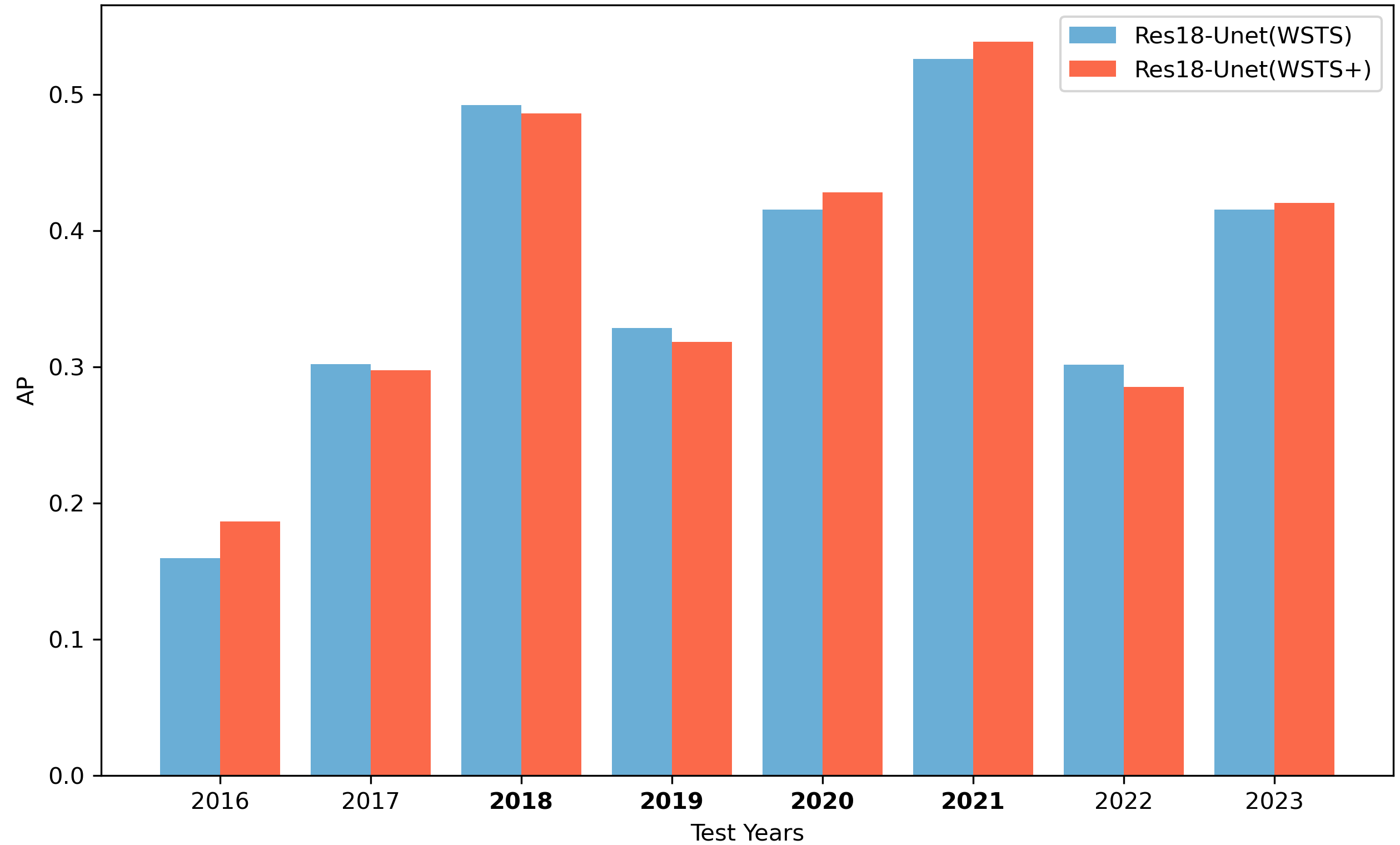}
    \caption{Performance breakdown by test year. Blue bars represent models trained on the original WSTS data, termed Res18-Unet(WSTS), while red bars represent those trained on WSTS+, termed Res18-Unet(WSTS+). The bolded x-axis ticks highlight original test years from WSTS. For Res18-Unet(WSTS+), we stratify its performance by year. For Res18-Unet(WSTS), we stratify by year to obtain performance for 2018 to 2021.  To obtain performance on the remaining years, we select the cross-validation fold with the best-performing model (as judged by its test fold error) and report its performance on the newly added WSTS+ years. }
    \label{fig:wsts-plot}
\end{figure}

Using our updated cross-validation scheme, we train our best $T=1$ models and our best $T=5$ model on WSTS+ and report the results in terms of mean average precision across all three feature sets in \cref{tab:WSTS+results}.  We see that the performance rank-order of our three models is still similar on WSTS+ as compared to WSTS. However, the overall performance is significantly lower for these models on WSTS+ as compared to WSTS (by roughly 0.1 AP). These results seem to contradict our initial hypothesis that additional training data would enable larger models and improve accuracy.  To investigate further, \cref{fig:wsts-plot} reports the \textit{per-year} performance for a Res18-Unet trained on either WSTS or WSTS+ (denoted Res18-Unet(WSTS) and Res18-Unet(WSTS+), respectively; see caption for details).  These results reveal that both models obtain very similar AP on every testing year, despite Res18-Unet(WSTS+) being trained on twice as many years of data in each fold as Res18-Unet(WSTS). Since the two models perform similarly across all years, the lower overall performance obtained on the WSTS+ benchmark in \cref{tab:WSTS+results} is likely due to the greater apparent difficulty of the new testing years (2016, 2017, 2022, and 2023), rather than lower predictive accuracy of the models trained on WSTS+.    

\begin{figure}
    \centering
    \includegraphics[width=0.85\linewidth]{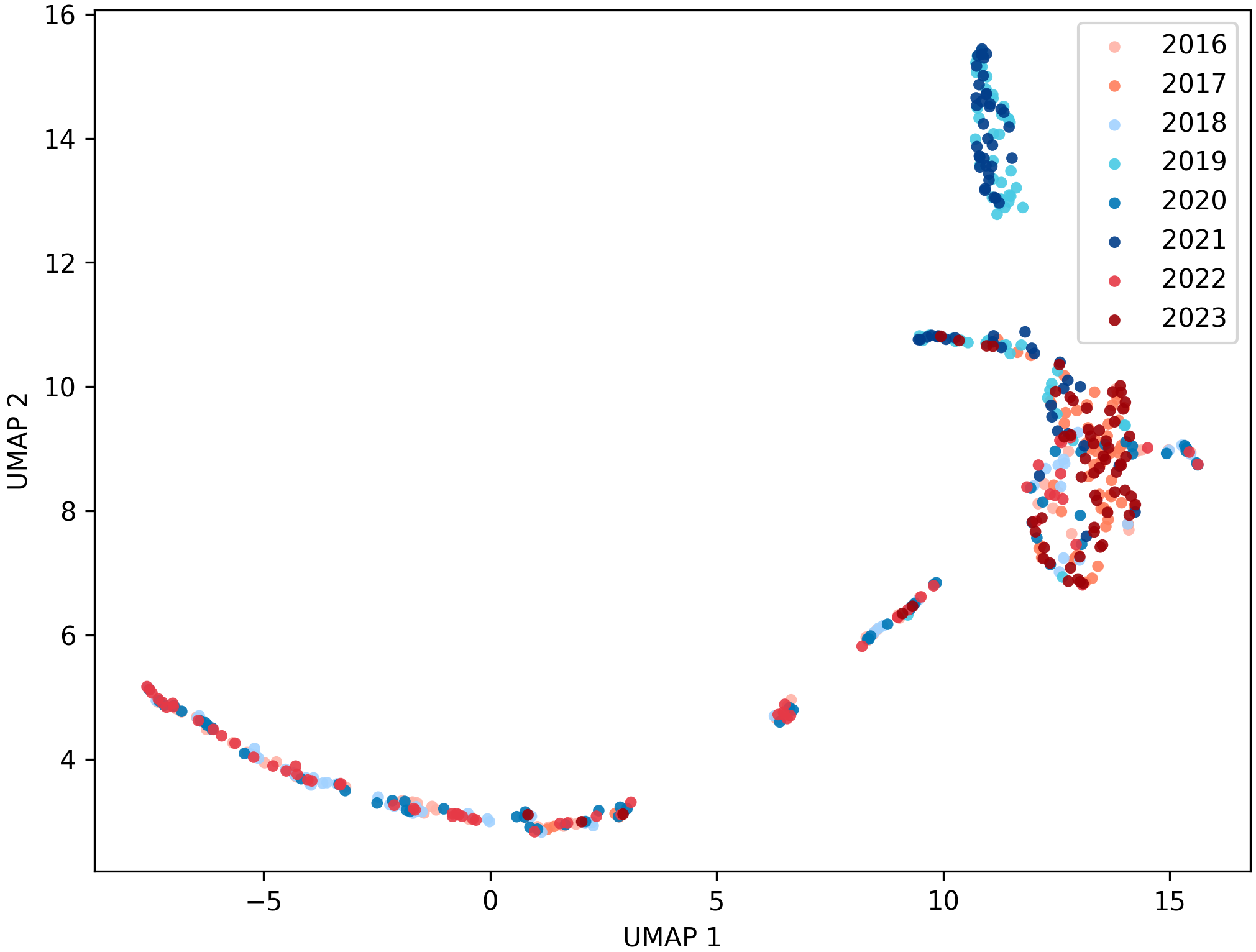}
    \caption{UMAP visualization of the input features across years. Each point represents the encoded features at the deepest layer of our best Res18-Unet encoder, with blue indicating original WSTS year and red newly added years in WSTS+}
    \label{fig:umap}
\end{figure}

\subsection{Domain Shift: A Potential Challenge to Scaling Data-Driven Wildfire Modeling}
\label{sub:domain_shift}
The results in \cref{fig:wsts-plot} raise a question: why does significantly increasing the quantity and diversity of the training data in WSTS+ lead to little or no improvement?  In \cref{sub:quality_assurance} of the supplement, we present evidence that, in pursuit of WSTS+, we accurately reproduced the preprocessing used for WSTS.  Therefore, we argue here that this result is likely caused by cross-year \textit{domain shift} \cite{quinonero-candela_dataset_2010}, wherein the joint probability distribution of the features and targets, denoted $p(x(t),y(t))$, varies across years.  There is a large literature about identifying and addressing domain shift (e.g., see \cite{quinonero-candela_dataset_2010,pan2009survey}), and comprehensively addressing these problems is beyond the scope of this work.  However, we seek here to provide evidence that domain shift is present in historical wildfire datasets, provide an initial characterization of it, and discuss the implications of it.  There are many different types of domain shift based upon precisely how $p(x(t),y(t))$ changes from training to testing conditions (or across years in our case) \cite{quinonero-candela_dataset_2010,kouw_introduction_2019, kull_patterns_nodate}).  We consider here two widely-studied types of shift: concept and covariate shift.  

\textit{Concept Shift} refers to changes in the conditional distribution $p(y(t)|x(t))$, which in a regression context \textit{generally} implies that the true underlying function $y(t)=f(x(t))$ is changing.  Under this hypothesis, we would expect that combining multiple years of data would likely lead to significant reductions in accuracy, since each year exhibits a different underlying relationship.  Our results in \cref{fig:wsts-plot} indicate that adding two additional years of training data, as is done in WSTS+, did not significantly impact accuracy, suggesting that significant concept shifts are unlikely.  In \cref{tab:cross_year_scores} we also report the results of an experiment where we train eight Res18-Unet models - one on data from each year - and then test each model on a disjoint test set from each year (see \cref{sup:cross_year} of the supplement for experimental details). The results indicate that, given a specific single testing year, most models achieve relatively similar accuracy, also suggesting they are each learning similar concepts.  

\textit{Covariate Shift} refers to change in the marginal distribution, $p(x(t))$.  Covariate shifts are thought to often have limited negative impact for high-capacity models \cite{quinonero-candela_dataset_2010}, such as our DNNs here.  Under this hypothesis, additional years of training data may be either beneficial or neutral, but not especially detrimental.  This hypothesis is therefore consistent with \cref{fig:wsts-plot}.  It is also corroborated by the results of \cref{tab:cross_year_scores} if we note that, for a specific testing year, the accuracy of most single-year models is similar to that obtained by WSTS models (trained on two years) and WSTS+ models (trained on four years) in \cref{fig:wsts-plot}. For example, if we take 2018 as the testing year, then the average of single-year models in \cref{tab:cross_year_scores} is nearly the same as the WSTS and WSTS+ models in \cref{fig:wsts-plot}.  In other words, despite significant differences in the years included in, and total size of the training data, these models usually perform similarly.  Notably, this also indicates that the WSTS benchmark did not benefit from additional training data: the WSTS models in \cref{fig:wsts-plot} do not perform differently (on average) than the single-year models in \cref{tab:cross_year_scores}.  As additional quantitative evidence, \cref{app:domain_shift} in the supplement provides substantial additional evidence that there is significant inter-year variability in environmental conditions (e.g., landcover composition, vegetation indices, and weather variables) and fire size and behavior.  This is summarized and corroborated by \cref{fig:umap}, which presents a UMAP visualization of the features extracted by our Res18-Unet for each year in WSTS+.  The results show that there is significant overlap in the feature distributions, but there are also significant apparent shifts across years.   


\textit{Conclusions} It is well-known within the fire science community that fire spread is impacted by a diverse set of environmental factors (e.g., weather, topography, and fuel)  \cite{rothermel1972mathematical, holsinger2016weather}, and these factors vary substantially across space and time.  For example, fire behavior experts have long established that annual weather patterns strongly influence both fire prevalence and extent  \cite{swetnam1990fire}.  Most of these important environmental factors are encoded by one or multiple input features in the WSTS dataset, providing a plausible physical basis for inter-year covariate shift.  Our analysis above provides substantial additional evidence that there is significant inter-year covariate shift in historical wildfire data, which would explain the limited benefits of additional training data in WSTS+.   
   
\begin{table}[h]
\centering
\caption{Cross-year results: Rows show the year the model was trained on, while columns show the year the model was tested on.}
\resizebox{\columnwidth}{!}{
\begin{tabular}{lccccccccc}

\toprule
Year & 2016    & 2017    & 2018    & 2019    & 2020    & 2021    & 2022    & 2023    & Avg     \\
\midrule
2016                      & \textbf{0.350} & 0.291 & 0.490 & 0.276 & 0.173 & 0.544 & 0.268 & 0.416 & 0.351 \\
2017                      & 0.242 & \textbf{0.300}  & 0.487 & 0.288 & 0.180 & 0.568 & 0.301 & 0.437 & 0.351 \\
2018                      & 0.265 & 0.297 & \textbf{0.576} & 0.313 & 0.194 & 0.595 & 0.344 & 0.465 & 0.381 \\
2019                      & 0.219 & 0.259 & 0.455  & \textbf{0.329} & 0.159 & 0.530 & 0.324 & 0.428 & 0.338  \\
2020                      & 0.222 & 0.263 & 0.501 & 0.285  & \textbf{0.220} & 0.572 & 0.295  & 0.460 & 0.352 \\
2021                      & 0.253 & 0.321 & 0.534  & 0.330 & 0.187 & \textbf{0.649} & 0.328 & 0.465 & 0.384 \\
2022                      & 0.227 & 0.249  & 0.460 & 0.261 & 0.163 & 0.508 & \textbf{0.390} & 0.416 & 0.334 \\
2023                      & 0.242 & 0.279 & 0.483 & 0.289 & 0.157 & 0.568 & 0.324 & \textbf{0.582} & 0.365 \\
\midrule
Avg                       & 0.253  & 0.282 & 0.498  & 0.296 & 0.179 & 0.567 & 0.322 & 0.459 & 0.357 \\
\bottomrule
\end{tabular}
}
\label{tab:cross_year_scores}
\end{table}

\section{Conclusion}
\label{sec:conclusion}
We investigated the problem of next-day wildfire spread prediction, systematically comparing a variety of (mostly) existing modeling strategies in two scenarios: single-day ($T=1$) and time-series ($T=5$) input, as illustrated in \cref{fig:problem_setting}. We conducted our experiments on the WSTS benchmark \cite{gerard2023wildfirespreadts} using a realistic 12-fold leave-one-year-out cross-validation and drew the following conclusions: 

\begin{itemize}
\item  Our study revealed which modeling strategies perform best, resulting in new models that obtain a $37\%$ and a $28 \%$ improvement, respectively, over the current WSTS state-of-the-art for single-day and time-series prediction. We find that substantial performance gains can be achieved not through novel architectures, but through the careful application and optimization of existing methods.

\item  A time-series model obtained the best overall performance, and time-series models usually outperformed comparable single-day models, suggesting time-series models are an important future area of research.

\item We introduce WSTS+, an extension of WSTS, that doubles the number of years of historical wildfire events in WSTS, and yields the largest existing public benchmark for \textit{time-series} spread prediction.

\item Analysis of WSTS and WSTS+ suggests that there is significant cross-year domain shift in historical wildfire data.  Preliminary investigation suggests it is primarily in the form of covariate shift, undermining the benefits of adding training data, but we hypothesize this problem may subside as total available hisorical data grows.   

\end{itemize}  
Future work may focus on investigating the nature of domain shift in historical wildfire data and overcoming any associated challenges, potentially enabling larger or more complex models (e.g., high capacity attention-based models) to realize their full potential performance.

{
    \small
    \bibliographystyle{ieeenat_fullname}
    \bibliography{main}
}
\clearpage
\setcounter{page}{1}
\maketitlesupplementary
\newcommand{\lt}{<}
\newcommand{\gt}{>}

\begin{figure}
    \centering
    \includegraphics[width=\linewidth]{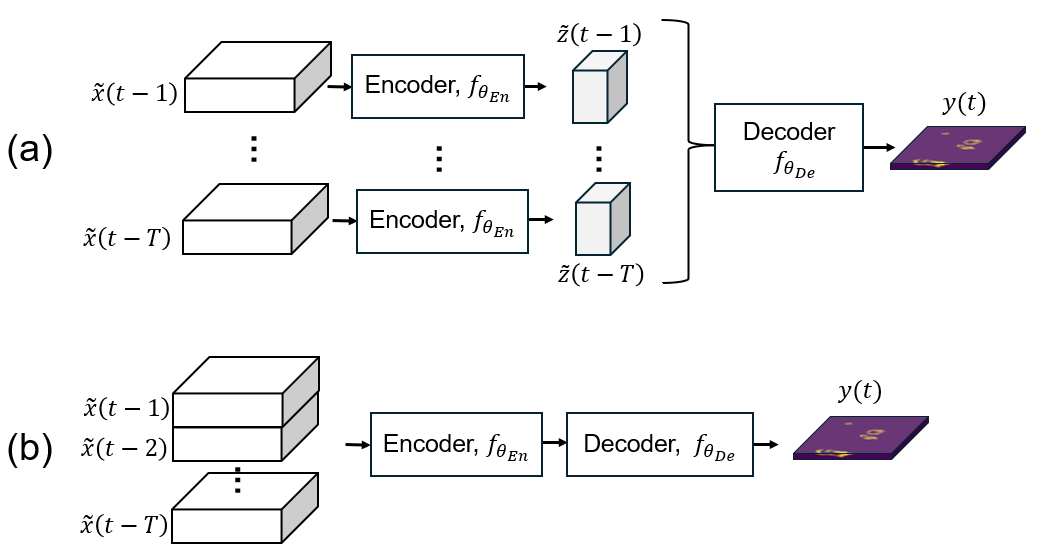}
    \caption{Illustration of (a) feature-level fusion, and (b) data-level fusion as we define it here. Further description is provided in the main text, and mathematical notation is described in \cref{sec:problem_setting}}
    \label{fig:data_vs_feature_fusion}
\end{figure}

\section{Experimental Details}
\label{app:exp_details}

\subsection{UTAE}\label{sub:utae} 
\begin{figure}
    \centering
    \includegraphics[width=0.85\linewidth]{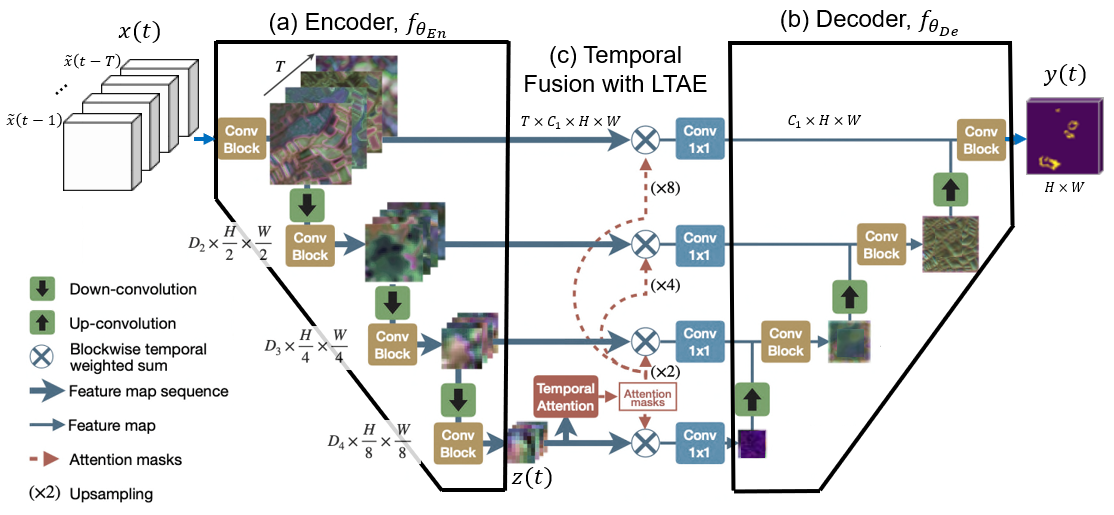}
    \caption{Illustration of the U-Net and UTAE models, adapted from \cite{garnot2021panoptic} to our wildfire problem: see description in main text.} 
    \label{fig:utae_and_unet_illustration}
\end{figure}

The UTAE \cite{garnot2021panoptic}, originally developed for satellite imagery is essentially a U-Net that has been modified to process a time-series of imagery, and was recently found successful for modeling wildfire spread \cite{gerard2023wildfirespreadts}.  We propose a novel modification of the time-series positional encodings and therefore discuss the technical details of the UTAE here. The UTAE encodes each entry in the time-series independently using a shared encoder shown in \cref{fig:utae_and_unet_illustration}(a), and then fuses the resulting embeddings from each day using a Lightweight Temporal Self-Attention (LTAE) block \cite{garnot2020lightweight}, shown in \cref{fig:utae_and_unet_illustration}(c). Given a $T$-length time-series of input, the encoder produces a series of $T$ embeddings $z(t)= \{ \tilde{z}(t-i) \}_{i=1}^{T}$ where $\tilde{z}(t) \in \mathbb{R}^{D_{4} \times \frac{H}{8} \times \frac{W}{8}}$ at the output of the last layer of the encoder.  Then the LTAE computes an attention mask, $a \in \mathbb{R}^{T \times \frac{H}{8} \times \frac{W}{8}}$, which is utilized to combine the $T$ embeddings. Before computing the temporal attention, LTAE adds a sinusoidal positional embedding, $p(\bar{t})$ to each input embedding, where $\bar{t} \in [1,365]$ is an integer representing the day of the year, and $p(\bar{t})$ maps $\bar{t}$ to a unique sinusoidal representation.  This positional embedding is motivated by the original application of UTAE to agricultural segmentation, where the appropriate segmentation depends heavily upon the day of the year. Once the attention mask is computed, it is then upsampled, and applied to the encoder embeddings output at each resolution to collapse the temporal dimension. After all temporal dimensions are collapsed, a conventional U-Net-like decoder is applied to the collapsed embeddings, as shown in \cref{fig:utae_and_unet_illustration}(b).  

\subsection{SwinUnet} SwinUnet \cite{cao2022swin} is a pure transformer-based Unet-shaped model that was first proposed for medical imagery segmentation. The model replaces the convolution blocks of the Unet with Swin Transformer blocks \cite{liu2021swin}, including them throughout the encoder, bottleneck, and decoder. They also rely on patch merging and patch expansion layers in the encoder and decoder, respectively, to downsample the input features and then upsample the extracted features and produce the segmentation mask. Finally, they preserve skip connections to concatenate shallow and deep features. The SwinUnet outperformed the Unet \cite{ronneberger2015u}, ViT \cite{chen2021transunet}, Att-Unet \cite{oktay2018attention}, and TransUnet \cite{chen2021transunet} on two medical benchmark datasets, and was shown to outperform the Unet on wildfire prediction \cite{zhou2025comparative}. Its state-of-the-art performance, ability to learn both global and long-range dependencies, and use of the more efficient Swin blocks make it a good candidate for our task. Since the model was developed for RGB images, we modify the \texttt{in\_chans} parameter to take in the number of channels of our multi-modal inputs (Veg: 7, Multi: 33, All: 40) instead of 3.  

\subsection{SegFormer} SegFormer \cite{xie2021segformer} is a recent, efficient, Transformer-based model developed for semantic segmentation. Whereas Swin focuses on improving the encoder using Transformers, SegFormer improves both the encoder, using a hierarchical Transformer that does not require positional encodings, and the decoder, using a lightweight MLP that makes the model efficient. SegFormer achieved excellent performance on the ADE20K and Cityscapes semantic segmentation benchmarks, surpassing state-of-the-art models like DeeplabV3+ and SETR. Several papers used SegFormer in wildfire-related tasks, including \cite{ouadou2025semantic, fusioka2024sentinel, shahid2023forest, saxena2025deep, cambrin2022vision} and found it to outperform CNNs on burned area delineation. Since the model does not use positional encodings, it can be fine-tuned/tested on any resolution. Therefore, we finetune it on our dataset without padding our input images. To control for model complexity, we use the SegFormer-B2 model as it uses the closest number of model parameters (27.5M) to that of the SwinUnet (27.2M).

\subsection{Model pre-training}
\label{app:pretraining}
To evaluate the effect of pre-training on the SwinUnet model, we load the \texttt{swin-tiny-patch4-window7-224} weights from HuggingFace onto each of our Swin blocks. These weights correspond to a Swin Transformer trained on ImageNet at 224x224 resolution. We zero-pad our input images (128x128) to match the expected input dimensions and benefit from the pre-trained weights. As for the Unet models, we follow \cite{gerard2023wildfirespreadts} and use the \texttt{segmentation\_models\_pytorch} implementation, and set \texttt{encoder\_weights} to \texttt{imagenet}, which loads a model with ImageNet pre-trained weights. The UTAE pre-training uses the PASTIS weights, released with the original paper \cite{garnot2021panoptic}. We use the 4th fold checkpoint, as it was the one with the highest performance. Finally, we load the \texttt{mit-b2} weights from HuggingFace to use the SegFormer-B2 encoder fine-tuned on Imagenet-1k.

\subsection{Training details}
To train our models, we adopt the implementations shared by \cite{gerard2023wildfirespreadts}, which can be found \hyperlink{https://github.com/SebastianGer/WildfireSpreadTS/tree/main/src}{in this GitHub repository}. The implementation relies on PyTorch Lightning for model creation, training, and testing and Weights \& Biases for model logging and metric visualization. All our models use a fixed batch size of 64, the AdamW optimizer, and a fixed optimized learning rate, as described in \cref{sec:experiments}. Also, following \cite{gerard2023wildfirespreadts}, we train our models for 10,000 iterations. Increasing the number of iterations to 15,000 and 20,000 did not yield any notable increases in performance.  For all runs in \cref{tab:results}, we report the mean test AP averaged over the 12 folds, and the standard deviation. During the hyperparameter search, we only use a single data fold (id = 2), train for 50 epochs, and pick the combination that yields the highest validation AP.

\section{Additional Analyses}
\subsection{Deployment Characteristics}
\label{sub:deployment_info}
In \cref{tab:deployment}, we provide key deployment characteristics for each model used in our benchmark. Parameter count refers to the total number of trainable parameters in Millions. We compute inference time by doing 10 warmup runs to stabilize the GPUs, then 100 inference runs, and report the mean in milliseconds $\pm$ standard deviation. GPU Memory Usage tracks peak GPU memory consumption in MB. We use \texttt{torchprofile.profile\_macs} to estimate total FLOPs (floating point operations). Training Time Estimation (in hours) simulates 20 forward and backward passes, then times a full training step, and extrapolates it to the full training regime (100 epochs using 1000 steps). Model size refers to the model weights file size in MB. 

The results in \cref{tab:deployment} show that the Res18-UNet offers the best balance, being a small (14M parameters), fast (2.5 ms inference), low-memory (55 Mb) model, resulting in excellent test AP (0.455). The Res50-UNet offers slightly higher accuracy (0.457) at the cost of double the amount of parameters, inference time, training time, and size. Both UNet-based models are relatively cheap computationally (1.8 and 3.1G FLOPs, respectively) and use a manageable amount of GPU Memory (70 MB and 375 MB, respectively), making them easier to deploy on machines with resource-constrained GPUs. 

The transformer-based models SwinUnet and SegFormer are slower (9-13 ms for inference and 1.8-2.0 h for training), and computationally heavier (3.7-6.1G FLOPs and 526-865 MB of GPU memory usage), yet without any AP gains. Finally, while the UTAE is compact in storage (4 MB only), it is very computationally expensive (10.6G FLOPs and 997 MB GPU memory usage) despite having the smallest number of parameters (1M). This is likely due to the expensive operations inside the temporal attention block (LTAE). Compared to the other models, it is rather slow in inference (9.5 ms) yet relatively fast in training (1 h). As such, it seems that the Res18-UNet is most optimal if deployment efficiency is the priority. Although SwinUnet and SegFormer don’t outperform the UNets in this setup, they may generalize better in other domains. UTAE offers a mix of fast training and lightweight model size with heavy computation and GPU memory usage.

\begin{table*}[h]
\centering
\caption{Model deployment characteristics and performance trade-offs}
\label{tab:deployment}
\begin{tabular}{lccccccc}
\toprule
\textbf{Model} & \textbf{Params (M)} & \textbf{FLOPs (G)} & \textbf{Inference (ms)} & \textbf{GPU Mem (MB)} & \textbf{Size (MB)} & \textbf{Training (h)} & \textbf{Test AP} \\
\midrule
Res18-UNet & 14.3 & 1.8 & 2.5±0.0  & 70  & 55  & 0.4 & 0.455 \\
Res50-UNet & 32.6 & 3.1 & 5.1±0.1  & 375 & 125 & 1.1 & 0.457 \\
SwinUnet   & 27.2 & 6.1 & 8.9±0.0  & 526 & 106 & 1.8 & 0.432 \\
SegFormer  & 27.5 & 3.7 & 12.7±0.8 & 865 & 105 & 2.0 & 0.448 \\
UTAE       & 1.1  & 10.6 & 9.5±1.0 & 997 & 4   & 1.0 & 0.452 \\

\bottomrule
\end{tabular}%
\end{table*}

\subsection{Why do simpler models outperform more complex ones?}
\label{sub:simpler_outperform_complex} 
In \cref{tab:results}, we found that the simpler convolution-based Res18-Unet outperformed its more complex, Transformer-based counterparts (SwinUnet and SegFormer). We hypothesized that this may be due to the realistic 12-fold leave-one-year-out (LOYO) cross-validation scheme adopted by the WSTS benchmark, penalizing the complex models for overfitting to temporal shifts. To test this hypothesis, we retrain our Res18-Unet, SwinUnet, and SegFormer models using a random 4-fold cross-validation scheme across fire events (i.e., each fire event, and all associated training instances only appear in one fold), and we report the results in \cref{tab:loyo_vs_random}. We find that performance (in AP) increases significantly when using event-based cross-validation ("random" in \cref{tab:results}) instead of LOYO validation, as expected.  However, the rank-order of the models remains unchanged, with Res18-Unet still outperforming the transformer-based SwinUnet and SegFormer, and by a similar margin. These results suggest that the cross-validation scheme is not responsible for the lower performance of more complex (e.g., transformer-based) models.  Furthermore, we find no evidence of overfitting among the transformer-based models in either LOYO or event-based "Random" cross-validation.  Therefore, it does not appear that overfitting is the cause of their inferiority compared to the ResUnet, and it (tentatively) appears that simpler convolutional models, such as the Res18-Unet may be generally superior for this task, although this is only a hypothesis and further study is needed to conclude.

\begin{table}[h]
    \centering
    \caption{Mean test AP $\pm$ standard deviation using vegetation features only (Veg), vegetation, land cover, topography, and weather (Multi), and All features, when training with 1 input day using the original leave-one-year-out (LOYO) 12-fold cross-validation scheme versus a random 4-fold cross-validation scheme. }
    \resizebox{\columnwidth}{!}{
    \begin{tabular}{lcccccc}
        \toprule
         & \textbf{Model} &  \textbf{X-val}  & \textbf{Veg} & \textbf{Multi} & \textbf{All} & \textbf{Params} \\
        \midrule
        & \multirow{2}{*}{Res18-Unet} & LOYO  & ${0.455 \pm 0.090}$ & $0.468 \pm 0.087$ & $0.460 \pm 0.084$ & \multirow{2}{*}{14.3M} \\
        &  & Random & $0.527 \pm 0.056$ & $0.540 \pm 0.058$ & $0.542 \pm 0.066$ &  \\
        & \multirow{2}{*}{SwinUnet} & LOYO  & $0.432 \pm 0.088$ & $0.437 \pm 0.082$ & $0.424 \pm 0.090$ & \multirow{2}{*}{27.2M} \\
        &  & Random & $0.493 \pm 0.127$ & $0.529 \pm 0.113$ & $ 0.511 \pm 0.090$ & \\
        & \multirow{2}{*}{SegFormer} & LOYO  & $0.433 \pm 0.080$ & $0.436 \pm 0.083$ & $0.423 \pm 0.087$ & \multirow{2}{*}{27.5M} \\
        &  & Random & $0.503 \pm 0.053$ & $0.515 \pm 0.046$ & $0.511 \pm 0.069$ &  \\
        \bottomrule
    \end{tabular}
    }
    \label{tab:loyo_vs_random}
\end{table}

\subsection{Statistical Significance} 
\label{sub:stat_significance}
\begin{figure}
    \centering
    \includegraphics[width=\linewidth]{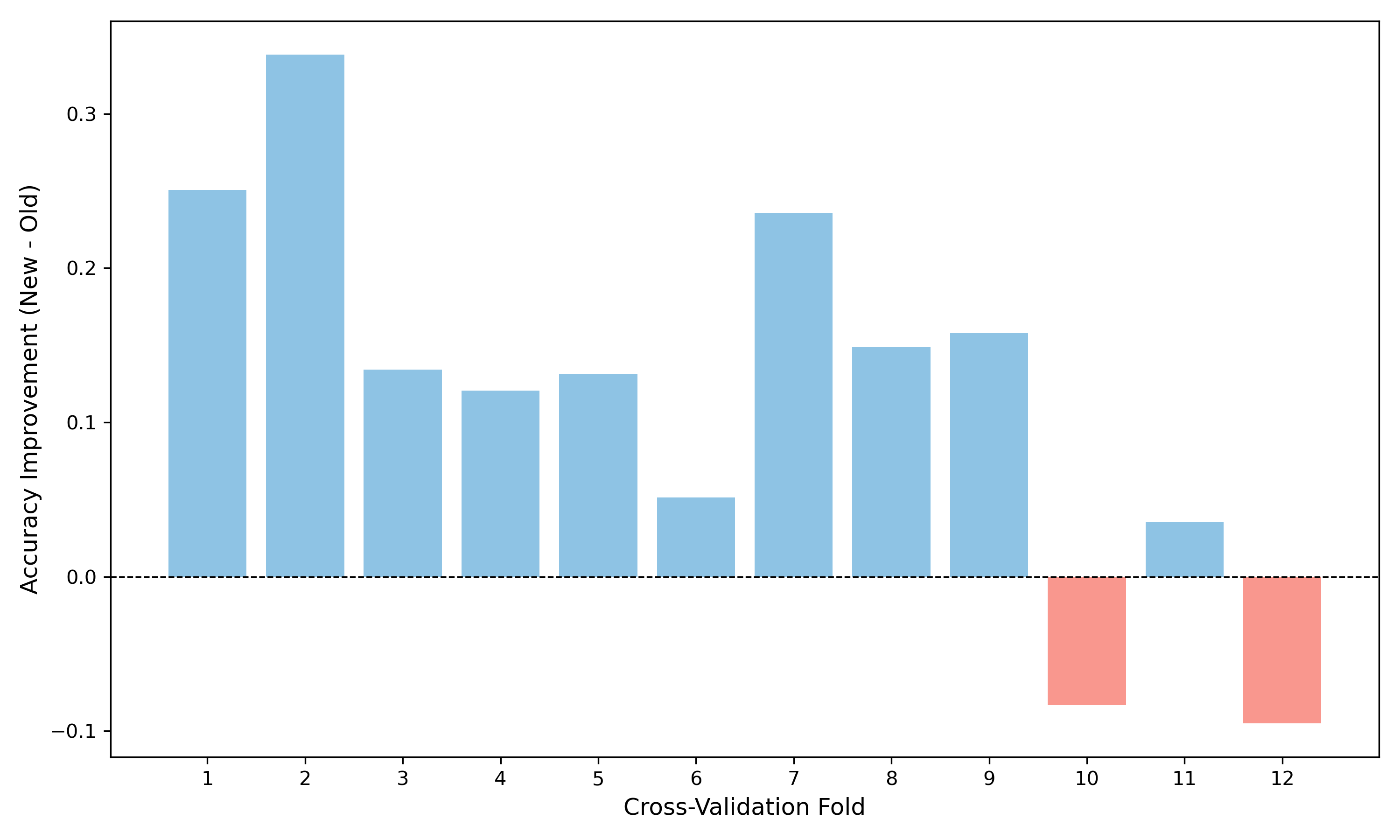}
    \caption{Per-Fold Accuracy Improvement (Res18-Unet (ours) vs. Original)}
    \label{fig:accuracy_improvement}
\end{figure}

To determine if the performance increase of our models was statistically significant with respect to the variance introduced by randomness in the training, validation, and testing data sets, we conducted a Wilcoxon signed-rank test on the twelve accuracy scores obtained from the 12 cross-validation folds of our best Res18-Unet, compared to the original Res18-Unet from \cite{gerard2023wildfirespreadts}. Given the distribution of the models' performance values (given by mAP) is highly non-Gaussian, we use the Wilcoxon test which does not assume a Gaussian distribution of outcomes, and tests whether the median differences are zero. The results indicate that our Res18-Unet model's mAP was statistically significantly higher than the previous model (W = 7.0, p = 0.0093). The relatively small W score indicates that the ranking differences between the models are consistently in favor of our model. Moreover, the p-value is below the 0.05 threshold, which supports the rejection of the null hypothesis. \cref{fig:accuracy_improvement} shows the per-fold accuracy improvement, with our model consistently achieving better average precision in 10 out of the 12 folds, further confirming the validity of our improvements.

\subsection{Failure Case Analysis}
\label{sub:failure_cases}
We present in \cref{fig:success_2018}, \cref{fig:success_2019}, \cref{fig:success_2020}, \cref{fig:success_2021} the 10 best predictions (as determined by AP) made by our best model, the Res18-Unet, for each testing year. On the other hand, \cref{fig:failure_2018}, \cref{fig:failure_2019}, \cref{fig:failure_2020}, \cref{fig:failure_2021} show the 10 worst predictions.  

Looking at the predictions, we observe that the model consistently fails (AP around 0.001) when the fire to predict is either extremely small (a few pixels), nonexistent in the previous day (a new, ignited fire), or considerably displaced relative to the current day (a spotting event, which is known to be a modeling challenging for the fire spread community \cite{trucchia2019randomfront}).
The model collapse is characterized by prediction maps being dominated almost entirely by false positives (shown in red). This is somewhat expected, as the model is not trained to predict the start of new fires. Moreover, we note that this behavior is consistent across all four test years, suggesting that these specific failures are due to the nature of the fire events, and not to data shift between years.  

Conversely, the model does best when the fires are larger, have a more consolidated structure, and grow/shrink around the same vicinity. It learned to consistently predict the bulk of the fire spread correctly (shown in green), with a small amount of false positives (shown in red) or false negatives (shown in blue), mostly around the edges of the fire. Still, the model achieves high performance on these samples (AP $\gt$ 0.9; F1 $\gt$ 0.8), showing that it learned to accurately predict larger fires.

\subsection{Fire Size Impact}
\label{sub:fire_size_impact}
\begin{figure}
    \centering
    \includegraphics[width=\linewidth]{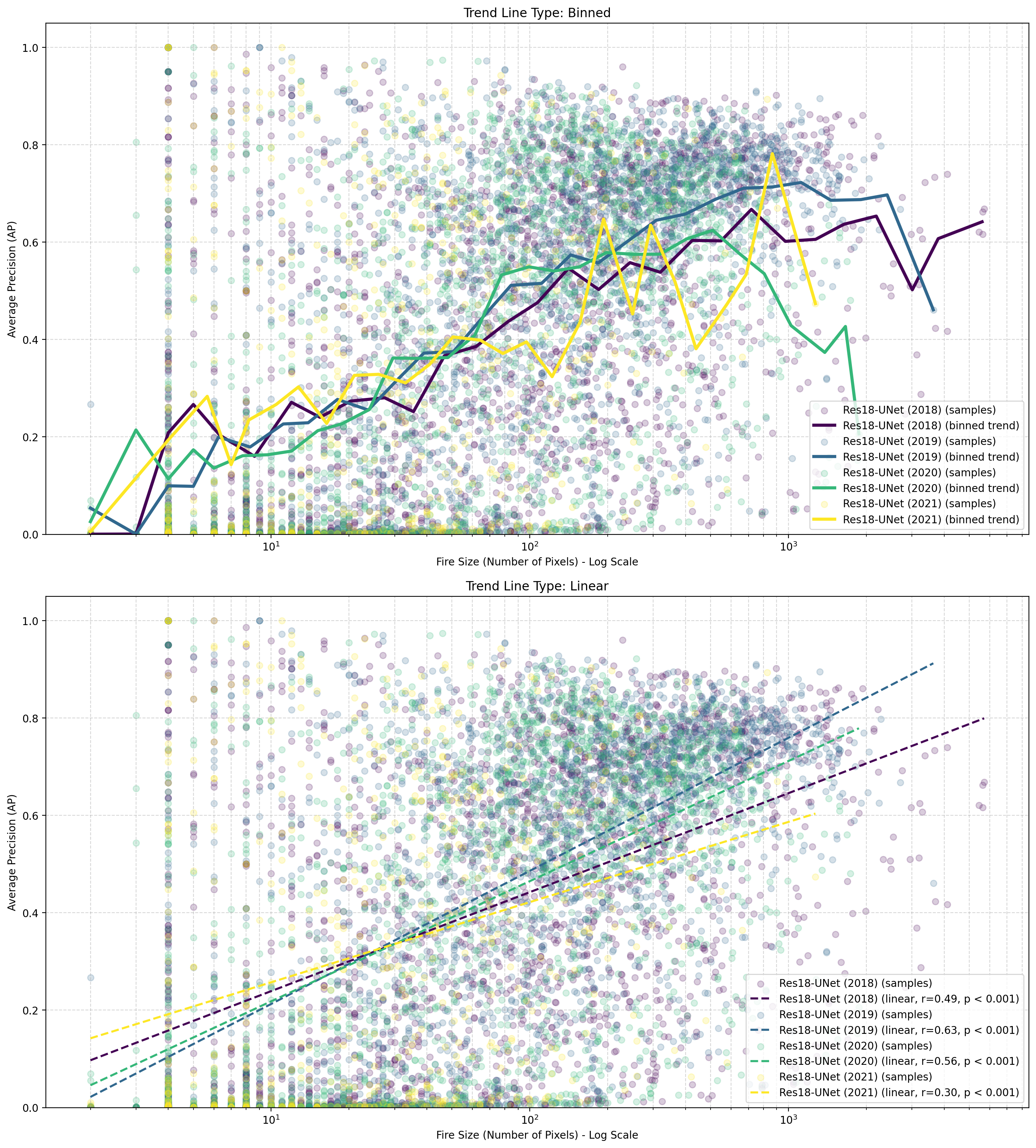}
    \caption{Scatters of test AP against fire size. Top: we overlay a line through binned averages. Bottom: we overlay a regression line.}
    \label{fig:ap_vs_fire_size}
\end{figure}

To further investigate the impact of fire size on the difficulty of prediction, we visualize in \cref{fig:ap_vs_fire_size} scatters of AP against fire mask size. We compute fire size as the total number of positive pixels in each ground truth mask, and compute the AP, per fire event, achieved by our best Res18-Unet, when tested on each year. Each dot represents a test instance, and in the top plot, we overlay a regression line between the fire size (on log scale) and the test AP. For the bottom plot, we first divide fire sizes into 30 bins and compute the mean AP for each bin, then plot a smoothed curve through the mean AP values.

Looking at the plots, we observe that fire sizes vary from very small ($\lt$ 10 pixels) to extremely large ($\gt$ 1000 pixels). We also note that the performance (measured by AP) is highly variable across scales. However, using smoothing allows us to confirm that our observations in \cref{sub:failure_cases} are not anecdotal but rather systematic: \textit{there exists a positive correlation between fire size and model performance across years, with larger fires being generally easier to predict}. 

In the top plot, we notice that the AP increases steadily with fire size up to a few hundred pixels, then plateaus. We also observe some differences between the years. For example, when the model is tested on 2019, it achieves the highest AP across medium-to-large fires. We also observe that the AP of the model tested on 2020 initially rises but reaches the lowest value of all models. The model tested on 2021 shows some fluctuation, with apparent instability in performance for large fires. Finally, the 2018 model seems to follow the smoothest curve.

In the bottom plot, we notice that the correlation strength varies throughout years, with 2019 (shown in blue) showing the strongest correlation (r = 0.63), and 2021 (shown in yellow) having the weakest correlation (r = 0.30). This means that the model is usually able to predict larger fires better, but this is inconsistent across years.

\begin{figure}
    \centering
    \includegraphics[height=\textheight]{figures/Res18-UNet__2018__best_predictions.png}
    \caption{10 best predictions made by the Res18-Unet on the 2018 test year.}
    \label{fig:success_2018}
\end{figure}

\begin{figure}
    \centering
    \includegraphics[height=\textheight]{figures/Res18-UNet__2018__worst_predictions.png}
    \caption{10 worst predictions made by the Res18-Unet on the 2018 test year.}
    \label{fig:failure_2018}
\end{figure}

\begin{figure}
    \centering
    \includegraphics[height=\textheight]{figures/Res18-UNet__2019__best_predictions.png}
    \caption{10 best predictions made by the Res18-Unet on the 2019 test year.}
    \label{fig:success_2019}
\end{figure}

\begin{figure}
    \centering
    \includegraphics[height=\textheight]{figures/Res18-UNet__2019__worst_predictions.png}
    \caption{10 worst predictions made by the Res18-Unet on the 2019 test year.}
    \label{fig:failure_2019}
\end{figure}

\begin{figure}
    \centering
    \includegraphics[height=\textheight]{figures/Res18-UNet__2020__best_predictions.png}
    \caption{10 best predictions made by the Res18-Unet on the 2020 test year.}
    \label{fig:success_2020}
\end{figure}

\begin{figure}
    \centering
    \includegraphics[height=\textheight]{figures/Res18-UNet__2020__worst_predictions.png}
    \caption{10 worst predictions made by the Res18-Unet on the 2020 test year.}
    \label{fig:failure_2020}
\end{figure}

\begin{figure}
    \centering
    \includegraphics[height=\textheight]{figures/Res18-UNet__2021__best_predictions.png}
    \caption{10 best predictions made by the Res18-Unet on the 2021 test year.}
    \label{fig:success_2021}
\end{figure}

\begin{figure}
    \centering
    \includegraphics[height=\textheight]{figures/Res18-UNet__2021__worst_predictions.png}
    \caption{10 worst predictions made by the Res18-Unet on the 2021 test year.}
    \label{fig:failure_2021}
\end{figure}

\section{WSTS+ Details}
\subsection{Collection Details}
\label{sub:collection_details}
To ensure our added wildfire events are most similar to the original ones, we follow the exact same collection procedure in \cite{gerard2023wildfirespreadts}. Namely, we rely on the Google Earth Engine script found \hyperlink{https://github.com/SebastianGer/WildfireSpreadTSCreateDataset}{in this repository}, to only collect wildfires that are larger than $10$ km$^2$, and we use the GlobFire dataset \cite{artes2019global} to identify wildfire events in the United States for 2016 and 2017. However, given GlobFire's temporal availability ends at 2021, we use the MTBS Burned Areas Boundaries Dataset \cite{MTBS_2017} to identify wildfires in 2022 and 2023.

\begin{figure}
    \centering
    \includegraphics[width=\linewidth]{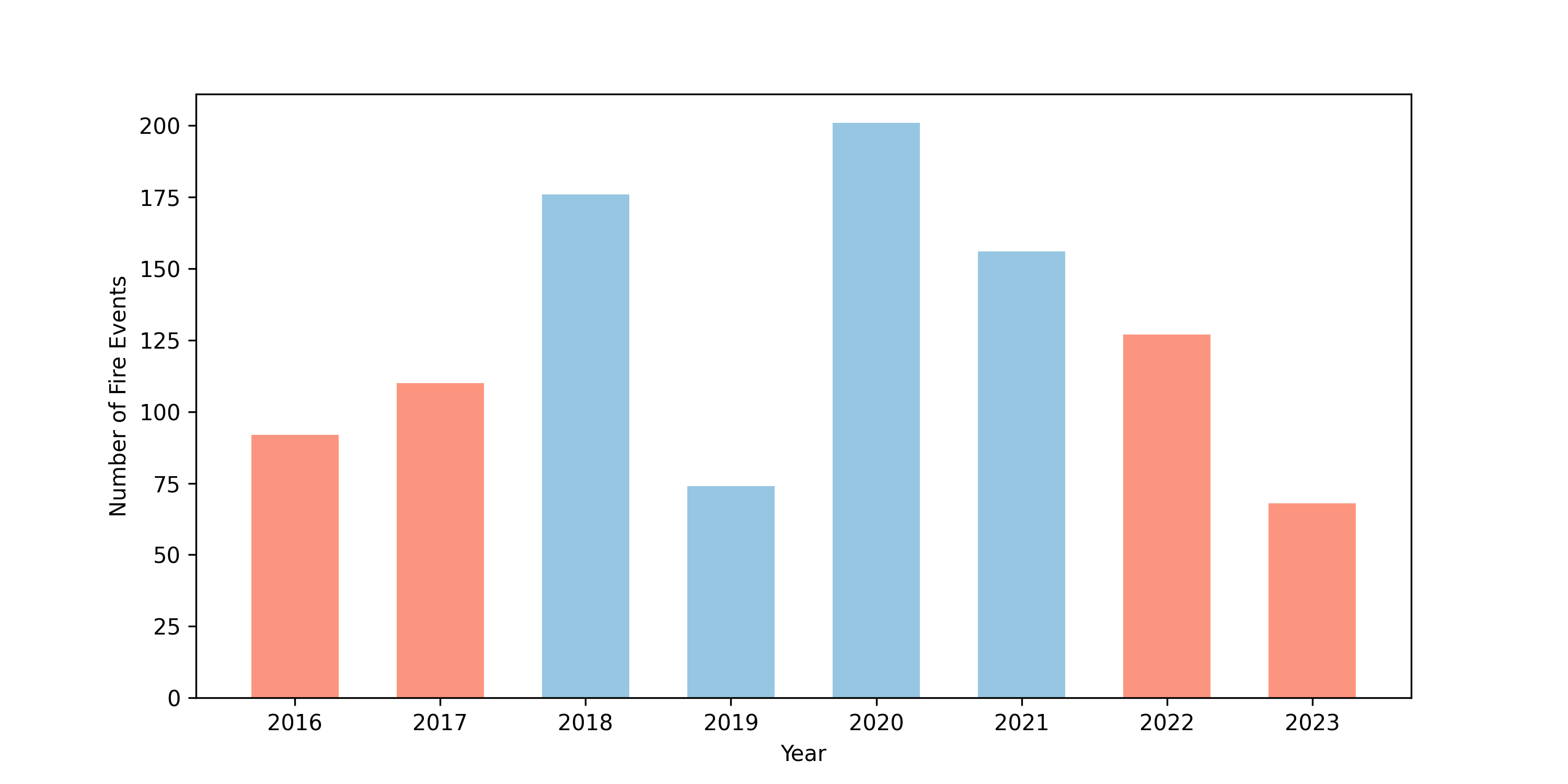}
    \caption{Distribution of fire events in WSTS+ per year}
    \label{fig:dist}
\end{figure}

The main differences between the datasets used for fire event \textit{identification} are that GlobFire relies on MODIS \cite{giglio2021modis} as a data source, which has a resolution of 500 meters, while MTBS uses Landsat imagery, which has 30 meter resolution. 
Furthermore, GlobFire returns burned area maps with start and end dates, while MTBS returns fire perimeters with start dates only. Regardless, we only use the centroid coordinate for both area maps and perimeters to download the fire masks. To account for the lack of fire end dates in MTBS, we collect 30 days of samples after the start date, with an additionnal buffer of 4 days before and after the fire events, similar to \cite{gerard2023wildfirespreadts}. We visualize the distribution of fire events in WSTS+ in \cref{fig:dist}. 




\subsection{Quality Assurance}
\label{sub:quality_assurance}
The new data were processed in the exact same way as the original WSTS data. To verify that it was done properly, we first replicated the downloading and processing of the original WSTS data (2018-2021), and measured the differences between our reproduction and the original data. We found that both are quantitatively similar. Specifically, we computed the mean pixel values of each data band for two folds (2018, 2019; and 2020, 2021) and found virtually no difference (max difference was 7.11e-04\% of each other). Further, to ensure these differences were not meaningful, we trained a Res18-Unet with T=1 on the Multi feature set (the best performing one from \cref{tab:results}) using our replicated WSTS and the original one. To verify that the results are similar, we show in \cref{tab:gerard-vs-jake} the test performance on each individual year and found that they are within a small numerical error of each other. 

Upon collecting the additional data in WSTS+, we computed the means and variances of each explanatory feature (e.g., wind speed, humidity, NDVI, EVI2, ERC) as well as the active fire feature across both original years (2018-2021) and newly added ones (2016, 2017, 2022, and 2023), and found that the distributions suggest some distribution shift. \cref{fig:kde_continuous}
shows kernel density estimates of the yearly distributions of multiple explanatory features in the dataset, highlighting varying degrees of cross-year domain shift across years. To validate this hypothesis, we conduct the cross-year experiments described in \cref{sup:cross_year}.

\begin{table}[h]
    \centering
    \caption{Comparison of model performance on WSTS original data versus our replicated WSTS data.}
    \begin{tabular}{lcc}
        \toprule
        \textbf{Test Year} & \textbf{Original} & \textbf{Replicated} \\
        \midrule
        2018 & 0.49533 & 0.48594 \\
        2019 & 0.31190 & 0.32115 \\
        2020 & 0.42248 & 0.41793 \\
        2021 & 0.56742 & 0.56031 \\
        \midrule
        Average & 0.44928 & 0.44633 \\
        \bottomrule
    \end{tabular}
    \label{tab:gerard-vs-jake}
\end{table}

\subsection{Inter-Year Domain Shift: Additional Analysis}
\label{app:domain_shift}

In this section, we perform additional analysis of domain shift.  It is well-known within the fire science community that fire spread is impacted by a diverse set of environmental factors such as weather (e.g., wind, temperature, precipitation), topography, and fuel quantity and type \cite{rothermel1972mathematical, holsinger2016weather}. These factors also vary substantially across locations and time. For example, fire behavior experts have long established that annual weather patterns strongly influence both fire prevalence and extent  \cite{swetnam1990fire}. 

Most of these important environmental factors are all encoded by one or multiple, input features in the WSTS dataset.  Furthermore, the creators of the WSTS dataset presented evidence that most of these features influence the likelihood of fire spread, to varying degrees (see Table 6 in \cite{gerard2023wildfirespreadts}).  Here, we build on those findings and report various evidence (e.g., visualizations, histograms, or statistics) that these features exhibit substantial inter-year variability as well, providing further evidence of cross-year domain shifts.  We focus our analyses on the features that were found to be (statistically) most influential of fire spread within the WSTS dataset, from the analysis in \cite{gerard2023wildfirespreadts}.   

\begin{figure}
    \centering
    \includegraphics[width=\linewidth]{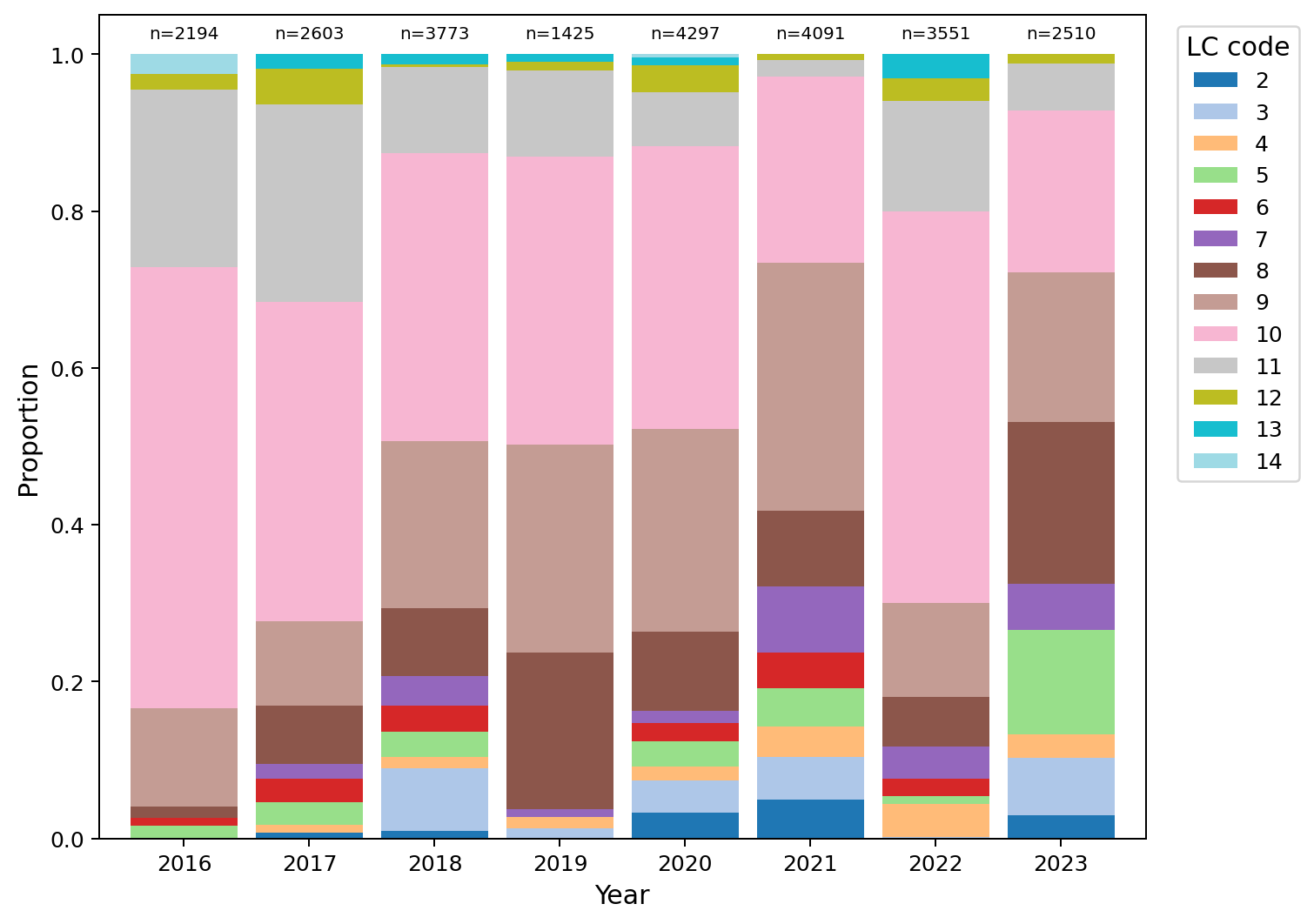}
    \caption{Histogram of landcover proportions across different years. }
    \label{fig:landcover}
\end{figure}

\paragraph{Landcover Classes }According to Table 6 in \cite{gerard2023wildfirespreadts}, the categorical variables with the highest importance (by absolute mean coefficient) were dominated by the different land cover classes (absolute coefficients between 5-28).  Looking at our plots, we notice in \cref{fig:landcover} that the proportions of landcover types vary substantially between years. For instance, LC 10 (Grasslands) was highly represented in 2016 (56.2\%) and 2022 (50.0\%) but comprised a much smaller proportion in 2021 (23.8\%) and 2023 (20.6\%). Additionally, LC 8 (Woody Savannas), shifts from being a minor component in 2016 (1.5\%) to a major landscape feature in 2019 and 2023 (~20\% of the area). We also notice a considerable decline for LC 11 (Permanent Wetlands), where it represented a significant portion of the landcover in 2016-2017 (+22\%), but shrunk to less than 6\% in more recent years like 2021 and 2023.

Furthermore, many landcover classes were nearly or totally absent in some years. For example, LC 2 (Evergreen Broadleaf Forests) is not represented in 2016, 2019, or 2022, and LC 14 (Cropland/Natural Vegetation Mosaics) is only found in 2016 (2.5\%) and 2020 (0.4\%). We also observe that the proportion of LC 5 (Mixed Forests) jumped to 13.3\% in 2023, after being absent in all prior years, suggesting a recent shift in the fire's environment.

\begin{figure}
    \centering
    \includegraphics[width=\linewidth]{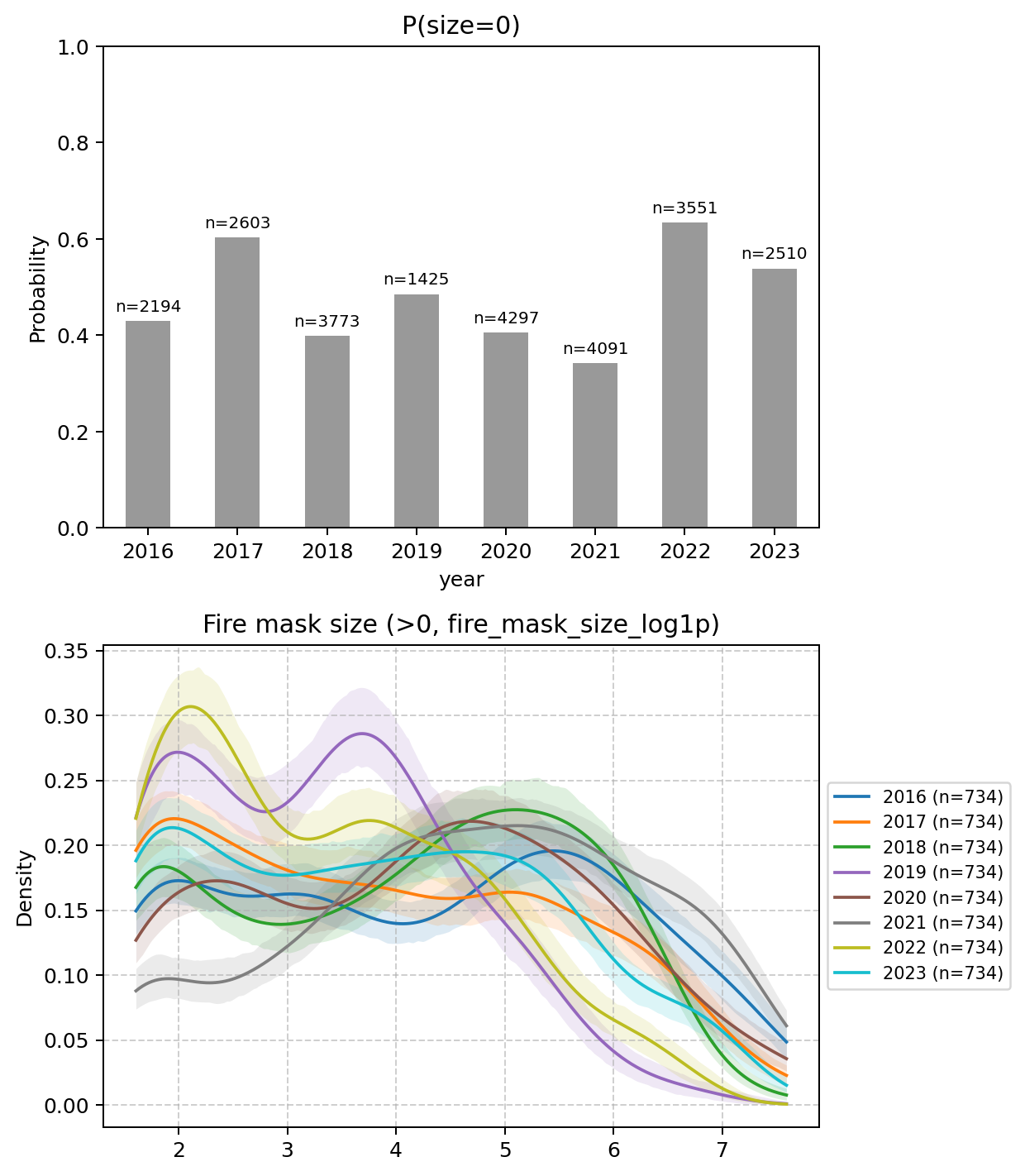}
    \caption{Top: Probability of a fire event having size zero for each year. Bottom: KDE showing distribution of fire sizes for fires that were larger than zero. The x-axis is log-transformed for better visualization, and the fire events balanced to ensure a fair comparison across years.}
    \label{fig:active_fire}
\end{figure}

\paragraph{Active Fire} Aside from landcover classes, active fire masks were -unsurprisingly- the features with the highest importance. As such, we visualize in \cref{fig:active_fire} the proportion of events where the fire mask size is zero for each year, alongside KDE plots showing fire size distribution after filtering out zero-sized events. To better visualize the skewed data, we log-transformed the x-axis, and to ensure a fair comparison, we balanced the number of fire events by randomly sampling events to match the smallest amount available for any given year. 

Similar to the landcover plots, \cref{fig:active_fire} reveals significant year-to-year variation, both for zero- and non-zero-sized events. For example, 2021 had the lowest    probability of zero-sized events (~35\%), i.e., a significantly higher proportion of observations in 2021 had active fires compared to other years. Looking back at the results in \cref{tab:cross_year_scores}, we observe that when 2021 is used as a testing year, we achieve the highest average AP, regardless of training year (0.567). On the other hand, 2017 and 2022 had much higher probabilities (+60\%), pointing to much higher fire inactivity, which exacerbates the class imabalance, and can also be seen in the relatively low results in \cref{tab:cross_year_scores}, where the average AP of models tested on these years was 0.282 and 0.322, respectively. 

Looking at the KDE plot, we further observe significant variability in the shape and typical size of fire events across years. For instance, the 2022 distribution (shown in olive) has a single, sharp peak at a smaller fire size, while the 2021 distribution (shown in purple) is much broader and has two distinct peaks, suggesting two common modes of fire size in that year. Both distributions are shifted to the left, indicating smaller fires on average. On the other hand, the gray curve for 2021 is shifted furthest to the right, meaning that larger fires were more common that year. Finally, 2016, 2017, 2018, and 2023 are quite mixed, with broad spread that indicates highly variable fire sizes.

\begin{figure*}
    \centering
    \includegraphics[width=\linewidth]{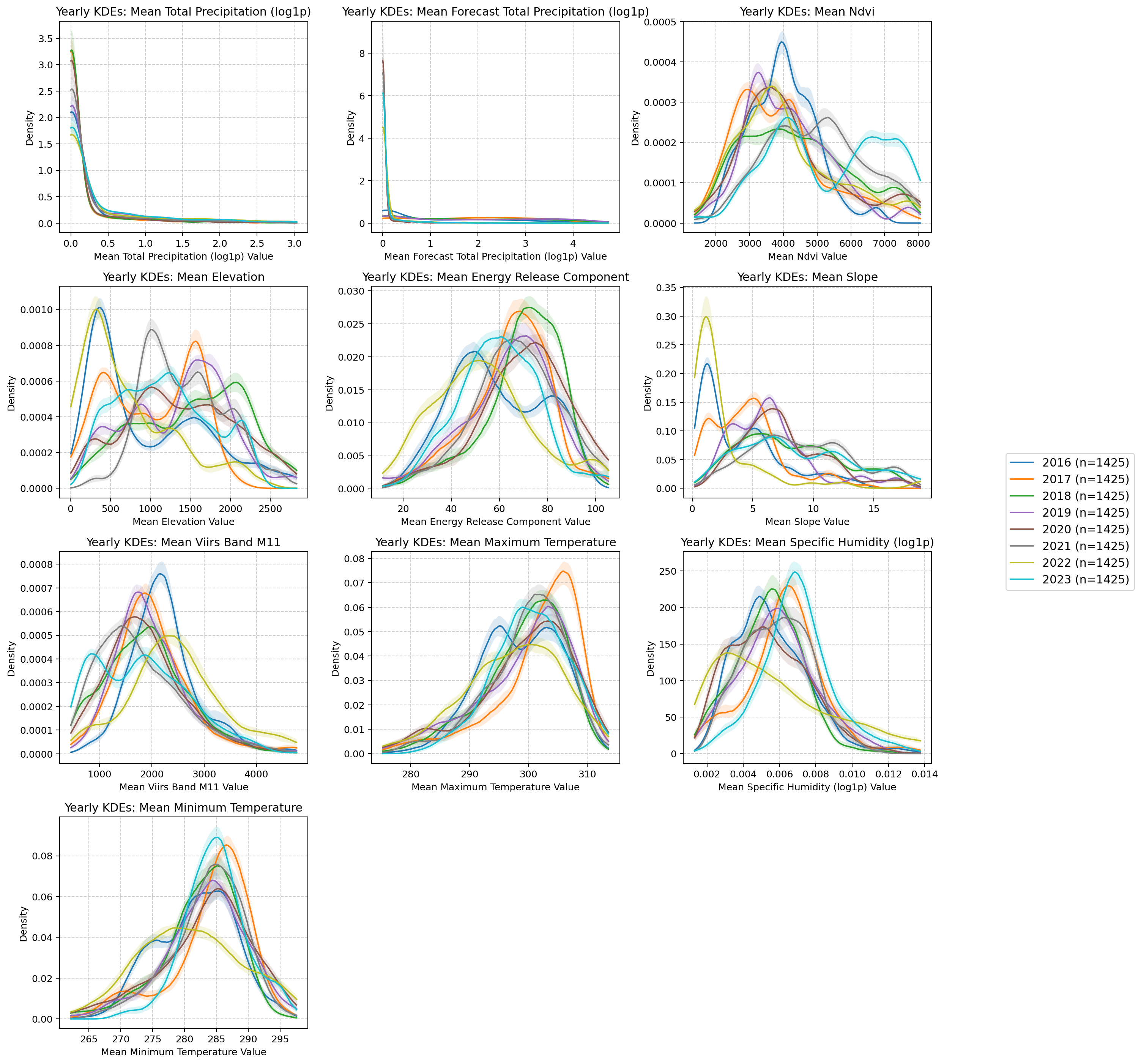}
    \caption{KDE facets of continuous features, plotted separately for each year.}
    \label{fig:kde_continuous}
\end{figure*}

\paragraph{Continuous Features} As for the continuous features, we find from \cite{gerard2023wildfirespreadts} that the ones with the highest importance were Total precipitation (-22.014), Forecast: Total precipitation (-9.865), NDVI (+4.178), Elevation (+2.933), Energy release component (+2.637), Slope (+2.406), VIIRS band M11 (+2.156), Maximum temperature (+0.948), Specific humidity (+0.857), Minimum temperature (+0.690). As such, we visualize KDE plots for each of them in \cref{fig:kde_continuous}. Similar to \cref{fig:active_fire}, we each line represents a KDE curve (with confidence intervals), showing the probability density across feature values in a given year. We also applied balanced subsampling, so that all years have the same number of samples, making comparisons fair. 

\begin{figure}
    \centering
    \includegraphics[width=\linewidth]{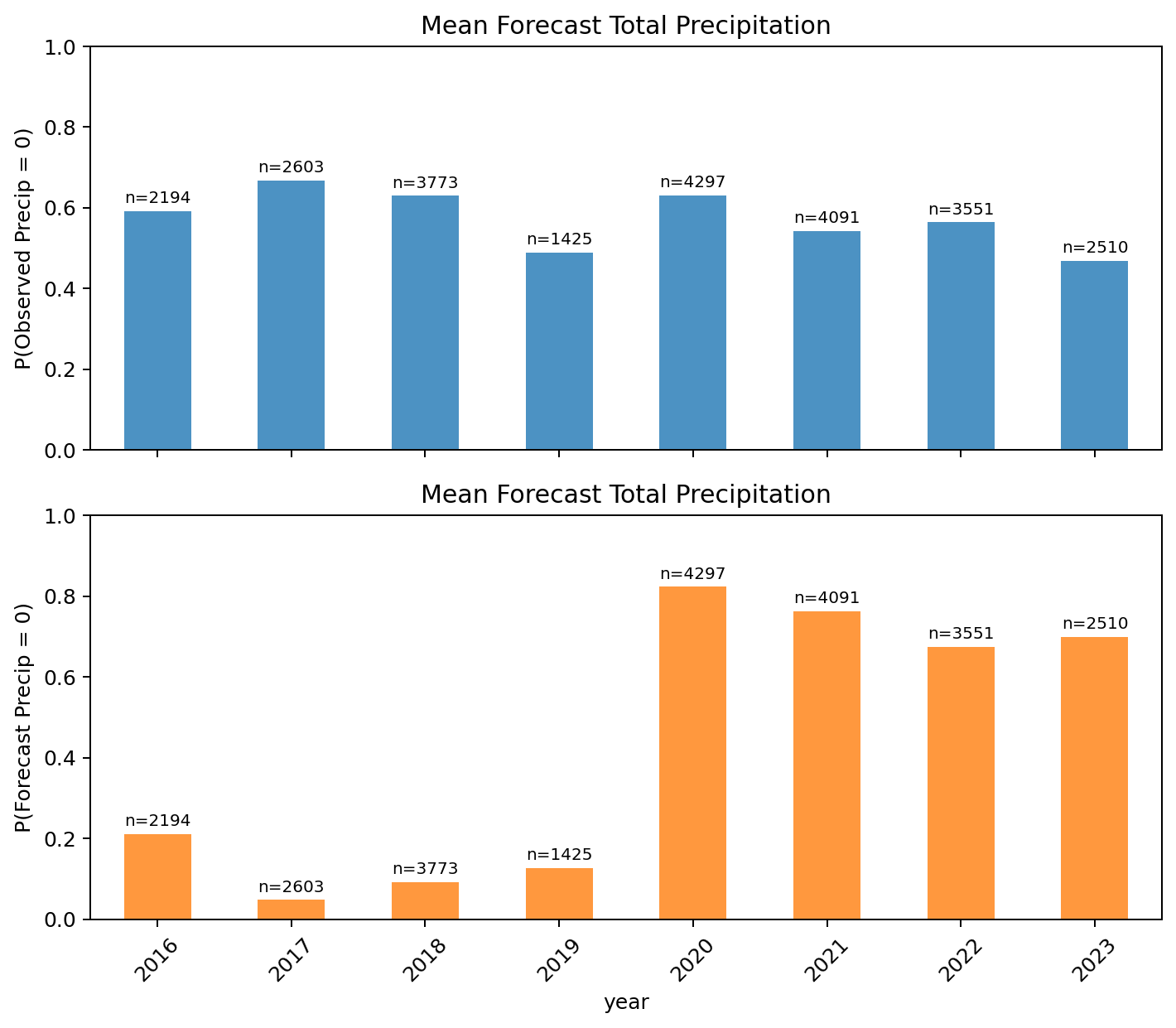}
    \caption{Distribution of forecast (top) and observed (bottom) no-rain events across years}
    \label{fig:precipitation}
\end{figure}

The precipitation plots are both similarly skewed toward very low values, indicating that the precipitation values do not vary as much from year to year. However, looking at the distribution of forecast zero-precipitation events in \cref{fig:precipitation} reveals a different insight: 2021 (~0.8) and 2022/2023 (~0.7) have many more zero-precipitation samples than 2016–2020, indicating that for some years, forecast no-rain events dominate, while in others forecasts predict more wet conditions. On the other hand, distributions of observed no-rain are nearly identical across years. Therefore, we can attribute the domain shift to the frequency of precipitation forecasts, but not to the differences in the magnitude of forecast or observed precipitation.

The remaining features show different levels of interannual variability, with each year's curve having a different shape (i.e., some are unimodal, some are bimodal), and peaking at different values. For instance, looking at the NDVI plot, we observe that 2016 is unimodal and peaks around an NDVI of 4000, while 2023 is bimodal and peaks much higher, around 7000, suggesting a greener year. This means that the type and condition of vegetation fueling the fires varied significantly from year to year. 

The elevation plot also shows significant variability of average elevation of fire locations between years, with the curves having completely different shapes and peaks. For instance, 2016 and 2022 peak around 400, while 2021 peaks at 1000, and 2018 at both 500 and 2000, suggesting that fires occurred in separate geographies with distinct elevations that year. 

Looking at the Energy Release Component (ERC) distributions reveals that curves shifted to the right, like 2018 (in green), experienced more severe drought conditions, therefore higher potential for intense fires. Overall, the significant spread across years highlights major differences in drought, a key driver of fire season severity. 

The slope plot shows that the 2022 fires (in olive) occurred on flatter terrain relative to the other years (low peaks). The M11, min/max temperatures, and specific humidity plots show significant overlap, with some outlier years (e.g., 2016 M11 peaking at 2000, correlating with hotter, intense fires; 2017 max temperature peaking at 305, indicating fire occurring in hotter conditions).

\begin{figure}
    \centering
    \includegraphics[width=\linewidth]{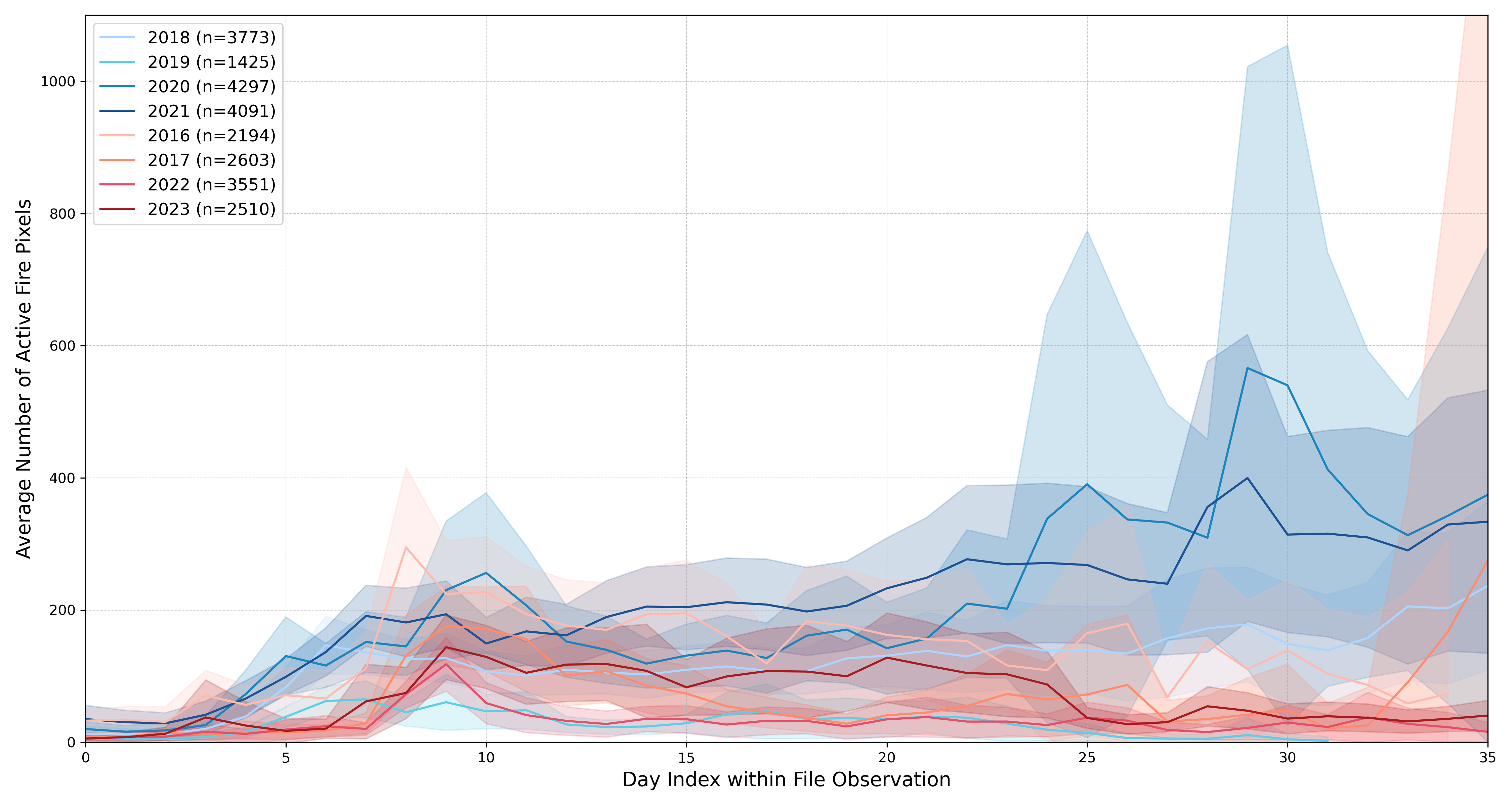}
    \caption{Comparing fire growth behavior for the different years in WSTS+ reveals significant interannual variability}
    \label{fig:analysis}
\end{figure}

\paragraph{Fire Growth}
To quantify the average daily fire growth pattern, we examine the number of active fire pixels on each day after ignition. In \cref{fig:analysis}, we plot the average progression of active fire pixels over the first 35 days following ignition, with a separate line for each year of data (blue lines represent WSTS years, while red one represent WSTS+ new years), and 95\% confidence intervals. As evidenced by the contrast in recorded fires between 2019 (1,422 fires) and 2020 (4,297 fires) shown in the legend, weather conditions drive substantial interannual fire variability. \cref{fig:analysis} highlights this variability between years, with many annual patterns falling entirely outside the confidence intervals of other years. Notably, years with greater fire activity, such as 2020 and 2021, exhibit more explosive growth during the first five days after ignition compared to other years.

\subsection{Cross-Year Experimental Design}
\label{sup:cross_year}
We discuss here the experimental design of the results shown in \cref{tab:cross_year_scores} in the main manuscript.  We trained a Res18Unet model on each training dataset listed in \cref{tab:dataset-splits}. Each year contributed a fixed quantity (and importance) of data samples (338 per year) to a shared validation set. We reached that number by reserving 20\% of the data of the year with the least amount of samples (2019 had 1351 total samples) as validation and used that number for all other years, resulting in 2704 validation samples across 8 years, which represented between  8.25\% and 16\% of the total samples of the remaining 7 years. The training sets contain $min(2000, |N|)$ where $N$ is the total data available for that year, after removing the validation samples. This ensured the training sets across years had roughly the same amount of data to train on (all years ended up having 2000 samples, except for 2016 and 2019, with 1751 and 1002 samples, respectively). We then evaluated each model’s average precision (AP) on each test set in \cref{tab:dataset-splits}. Notably, we ensured the training and validation sets for each year contain disjoint sets of fire events. 
\begin{table}[h]
    \centering
    \caption{Training, validation, and testing set sizes for each year, used for the cross-year train/testing.}
    \resizebox{\columnwidth}{!}{
    \begin{tabular}{lcccccccc}
        \toprule
        \textbf{Year} & \textbf{2016} & \textbf{2017} & \textbf{2018} & \textbf{2019} & \textbf{2020} & \textbf{2021} & \textbf{2022} & \textbf{2023} \\
        \midrule
        Training Set Size & 1385 & 1697 & 2000 & 692 & 2000 & 2000 & 2000 & 1818 \\
        Validation Set Size & \multicolumn{8}{c}{2704} \\
        Testing Set Size & 338 & 338 & 338 & 338 & 338 & 338 & 338 & 338 \\
        \bottomrule
    \end{tabular}
    }
    \label{tab:dataset-splits}
\end{table}


\end{document}